%% file: main.tex
\documentclass{article}
\usepackage[utf8]{inputenc}

\usepackage{geometry}
\usepackage{pdflscape}
\usepackage{graphicx}
\usepackage{enumitem}
\usepackage{amsmath}
\usepackage{amsthm}
\usepackage{amsfonts}
\usepackage[ruled]{algorithm2e}
\usepackage{hyperref}
\usepackage{cleveref}
\usepackage{caption}
\usepackage{subcaption}
\usepackage{xcolor}
\usepackage{bbm}
\usepackage{tcolorbox}
\hypersetup{
    colorlinks = true,
    linkbordercolor = {white},
    linkcolor={cyan},
    urlcolor={blue},
    citecolor= gray
}
\usepackage{commands}

\title{MiWaves Reinforcement Learning Algorithm}
\author{
Susobhan Ghosh$^1$
,
Yongyi Guo$^2$,
Pei-Yao Hung$^{3}$,
Lara Coughlin$^4$,
Erin Bonar$^4$, \\
Inbal Nahum-Shani$^3$,
Maureen Walton$^4$, and 
Susan Murphy$^1$\\
\small $^1$\textit{Department of Computer Science, Harvard University}\\
\small $^2$\textit{Department of Statistics, University of Wisconsin-Madison}\\
\small $^3$\textit{Institute for Social Research, University of Michigan}\\
\small $^4$\textit{Department of Psychiatry, University of Michigan}\\
\small susobhan\_ghosh@g.harvard.edu, guo98@wisc.edu, peiyaoh@umich.edu, \\
\small laraco@med.umich.edu, erinbona@med.umich.edu, inbal@umich.edu, \\
\small waltonma@med.umich.edu, samurphy@g.harvard.edu
}

\begin{document}
\date{}

\maketitle

\begin{abstract}
    The escalating prevalence of cannabis use poses a significant public health challenge globally. In the U.S., cannabis use is more prevalent among emerging adults (EAs) (ages 18-25) than any other age group, with legalization in the multiple states contributing to a public perception that cannabis is less risky than in prior decades. To address this growing concern, we developed MiWaves, a reinforcement learning (RL) algorithm designed to optimize the delivery of personalized intervention prompts to reduce cannabis use among EAs. MiWaves leverages domain expertise and prior data to tailor the likelihood of delivery of intervention messages. This paper presents a comprehensive overview of the algorithm's design, including key decisions and experimental outcomes. The finalized MiWaves RL algorithm was deployed in a clinical trial from March to May 2024.
\end{abstract}

\tableofcontents

\;\;

\input{intro}

\input{background}
\input{mixed_effects}
\input{simulator}
\input{rl_design}

\clearpage
\bibliographystyle{unsrtnat}
\bibliography{supp.bib}

\end{document}

%% file: intro.tex
\section{Introduction}
The escalating prevalence of cannabis use poses a significant public health challenge globally. In the U.S., cannabis use is more prevalent among emerging adults (EAs) (ages 18-25) than any other age group \cite{SAMHSA}, with legalization in the multiple states contributing to a public perception that cannabis is less risky than in prior decades \cite{carliner2017cannabis}. Although not all cannabis users experience harm, early onset of use is associated with significant physical and mental health consequences. To address this growing concern, we developed MiWaves, a reinforcement learning (RL) algorithm designed to optimize the delivery of personalized intervention prompts to reduce cannabis use among EAs. MiWaves leverages domain expertise and prior data to tailor the likelihood of delivery of intervention messages. 

This paper documents the development of the reinforcement learning (RL) algorithm for MiWaves. Reinforcement learning (RL) \cite{sutton2018reinforcement} is an area of machine learning where algorithms learn to select a sequence of decisions  in order to maximize an outcome. In MiWaves, the RL algorithm optimizes decisions regarding the delivery of engagement prompts to participants with to goal of maximizing their app engagement. The finalized MiWaves RL algorithm was deployed in a clinical trial (called the MiWaves clinical trial or the MiWaves pilot study) from March to May 2024.

In the MiWaves pilot study, the RL algorithm made decisions for $m = 122$ participants, where each participant was in the study for 30 days. The RL algorithm takes decisions twice daily for each participant in the study. For MiWaves, the decision points are immediately after the conclusion of the participant's morning and evening check-in windows, or immediately after the completion of their corresponding check-ins. 
The RL algorithm periodically updates a model for participants' outcome (app engagement) using the participant data collected by the app. Specifically for MiWaves, the RL algorithm has two types of updates - model parameter updates which occur daily at 4 AM EST, and hyper-parameter updates which occur once weekly on Sundays at 4 AM EST.

\subsection{Formulating the RL problem}
To formulate the RL problem, we first define a state, action, and reward. For each of the components, we use subscript $i$ to denote the participant and subscript $t$ to denote the decision point. The state $\state{t}{i}$ for participant $i$ at decision point $t$ describes the current context of the participant. We define algorithm state features as relevant features used by the algorithm that provide a summary of state $\state{t}{i}$. 
For example, state features could be the participant's recently reported cannabis use, or the time of day. The action $\action{t}{i}$ is the decision made by the RL algorithm for participant $i$ at decision point $t$. A policy is a function that takes in input state $\state{t}{i}$ and outputs action $\action{t}{i}$. For MiWaves, the action is a binary decision of whether to deliver an engagement prompt ($\action{t}{i}=1$) or not ($\action{t}{i}=0$).  The policy first calculates the  probability $\pii{t}{i}$ of selecting $\action{t}{i}=1$ and then uses $\pii{t}{i}$ to sample $\action{t}{i}$ from a Bernoulli distribution. That is, $\action{t}{i}$ is micro-randomized at time $t$ with  probability  $\pii{t}{i}$. For MiWaves, the policy is initially warm-started by using the prior distribution (Section~\ref{sec:priors}) to compute $\pii{t}{i}$ . The reward $\reward{t+1}{i}$ is a function of the proximal outcome, designed to  reward the algorithm for selecting good actions for participants.
In MiWaves, the reward is a function of the participant app engagement (proximal outcome). See Section~\ref{sec:rl_framework} for a detailed description of the states, actions and reward for MiWaves. 

We now describe how the RL algorithm makes decisions and learns throughout the study. The MiWaves RL algorithm is an \emph{online} RL algorithm which  interacts with participants, observing participant context, taking an action, and receiving some feedback or reward for its action in that given context. It continues to incorporate this feedback into its policy to learn about participants and improve the decision of which action to take in each context. Online RL generally has two major components: (i) the online learning algorithm (\cref{sec:online_learning_alg}); and (ii) the action-selection procedure (\cref{sec:act_select}). In MiWaves, at a given decision point $t$, the RL algorithm observes participant $i$'s observed state $\state{t}{i}$ and selects action $\action{t}{i}$ with  probability $\pii{t}{i}$ utilizing its action-selection procedure. After the RL algorithm selects $\action{t}{i}$, the algorithm then observes the reward $\reward{t+1}{i}$ from the participant. At update times, the RL algorithm updates the model of the participant environment using a history of states, actions, and rewards up to the most recent decision point, by utilizing it's online learning algorithm. 

\subsection{Code}
The code for the RL algorithm, along with the associated API service can be found \href{https://github.com/StatisticalReinforcementLearningLab/miwaves_rl_service}{here}.

%% file: background.tex
\subsection{Reinforcement Learning Framework}
\label{sec:rl_framework}
This section provides a brief overview of the Reinforcement Learning (RL) \cite{sutton2018reinforcement} framework used in this work with respect to the MiWaves study.



\begin{itemize}
    \item \textbf{Actions}: Binary action space, i.e. $\mathcal{A} = \{0, 1\}$ - to not send a MiWaves message ($0$) or to send a MiWaves message ($1$). 
    \item \textbf{Decision points}: $T = 60$ decision points per participant. The trial is set to run for $30$ days, and each day is supposed to have $2$ decision points per day. Therefore, we expect to have $60$ decision points per participant.
    \item \textbf{Reward}: Let us denote the reward for participant $i$ at decision time $t$ by $\reward{t+1}{i}$. One of the assumptions underpinning the MiWaves study is  that higher intervention engagement (self-monitoring activities and using suggestions by MiWaves to manage mood, identify alternative activities, plan goals, etc.)  will lead to  lower cannabis use. 
    Thus the reward for the RL algorithm is based on a combination of check-in completion and self-reports of use of MiWaves message suggestions.   
    This reward is defined as follows: 
    \begin{itemize}
        \item 0: did not open the app and did not complete the check-in (no app interaction)
        \item 1: opened the app and browsed it for more than 10 seconds outside of the check-in, but did not complete the check-in.
        \item 2: completed the check-in, and responded no to the \textit{Activity} question (whether the participant thought about or used any of the ideas from the MiWaves messages sent by the app)
        \item 3: completed the check-in, and responded yes to the \textit{Activity} question (whether the participant thought about or used any of the ideas from the MiWaves messages sent by the app)
    \end{itemize}
    \item \textbf{States}: Let us denote the state observation of the participant $i$ at decision time $t$ as $\state{t}{i}$. A given state $S = \{S_1, S_2, S_3\}$ is defined as a 3-tuple of the following binary variables (omitting the participant and time index for brevity):
    \begin{itemize}
        \item $S_1$: Recent intervention engagement - average of the past $X$ rewards ($X=3$). Categorized to be low (0) vs high (1). The binary state $S_1$ is then computed as follows: 
        \begin{align}
            S_1 = \left\{
            	\begin{array}{ll}
            		1  & \mbox{if } avg_{X}(R) \geq 2 \\
            		0 & \mbox{otherwise}
            	\end{array}
                \right.
            \label{eqn:S1}
        \end{align}
        At decision points $t=1$ and $t=2$, we take the average of the rewards observed upto that decision point.
        
        \item $S_2$: Time of day of the decision point - Morning (0) vs. Evening (1).

        \item $S_3$: Recent cannabis use (CU) reported in the check-in in the last $Y$ decision points ($Y=1$) (Used or not used). The binary state $S_3$ is then assigned as follows:
        \begin{align}
            S_3 = \left\{
            	\begin{array}{ll}
            		0  & \mbox{if } avg_{Y}(CU) > 0 \\
            		1 & \mbox{otherwise}
            	\end{array}
                \right.
            \label{eqn:S3}
        \end{align}

        At decision point $t=1$, since we have no reported cannabis use of the participant, we set $S_3$ to be 1. We do so because we expect the participants in the MiWaves study to be using cannabis regularly (at least 3 times a week).
    \end{itemize}
    Overall, we represent all the favorable states as $1$ (not using cannabis, engaging with the app), and the unfavorable values as $0$ (using cannabis, non-engaging with the app).
\end{itemize}

%% file: mixed_effects.tex
\section{Online Learning Algorithm}
\label{sec:online_learning_alg}
This section details the online learning algorithm - specifically the algorithm's reward approximating function and its model update procedure.


\subsection{Reward Approximating Function}
\label{rl:reward_model}
One of the key components of the online learning algorithm is its reward approximation function, through which it models the participant's reward. Recall that the reward function is the conditional mean of the reward given state and action. We chose a Bayesian Mixed Linear Model to model the reward. Linear models are well studied, and also easily interpretable by domain experts, allowing them to critique the model. Further, mixed models allow the RL algorithm  to adaptively pool and learn across participants while simultaneously personalizing actions for each participant.  

For a given participant $i$ at decision time $t$, the RL algorithm receives the reward $\reward{t+1}{i}$ after taking action $\action{t}{i}$ in the participant's  current state $\state{t}{i}$. Then, the reward model is written as:
\begin{align}
    \reward{t+1}{i}
    = g(\state{t}{i})^T \boldsymbol{\alpha_i}  + \action{t}{i} f(\state{t}{i})^T \boldsymbol{\beta_i} + \epsilon_i^{(t)}
\end{align}
where $\epsilon_i^{(t)}$ is the noise, assumed to be gaussian i.e. $\bs{\epsilon} \sim \mathcal{N}(\bs{0}, \sigma_{\epsilon}^2\bs{I}_{t m_t})$, and $m_t$ is the total number of participants who have been or are currently part of the study at time $t$. Also $\boldsymbol{\alpha_i}$, $\boldsymbol{\beta_i}$, and $\boldsymbol{\gamma_i}$ are weights that the algorithm wants to learn. $g(S)$ and $f(S)$ are functions of the algorithm state features (defined in \cref{sec:rl_framework}), defined as:
\begin{align}
    g(S) = [1, S_1, S_2, S_3, S_1 S_2, S_2 S_3, S_1 S_3, S_1 S_2 S_3]\\
    f(S) = [1, S_1, S_2, S_3, S_1 S_2, S_2 S_3, S_1 S_3, S_1 S_2 S_3]
\end{align}

Please refer to \Cref{sec:results} for the justification behind the design choice of these functions.

To enhance robustness to misspecification of the baseline, $g(\state{t}{i})^T \boldsymbol{\alpha_i} $, we utilize action-centering \cite{greenewald2017action} to learn an over-parameterized version of the above reward model:
\begin{align}
    \reward{t+1}{i}
    &= g(\state{t}{i})^T \boldsymbol{\alpha_i}  + (\action{t}{i} - \pii{t}{i}) f(\state{t}{i})^T \boldsymbol{\beta_i} + (\pi_i^{(t)})f(\state{t}{i})^T \boldsymbol{\gamma_i} + \epsilon_i^{(t)}
\end{align}
where $\pii{t}{i}$ is the probability of taking action $\action{t}{i} = 1$ in state $\state{t}{i}$ for participant $i$ at decision time $t$. We refer to the term $g(\state{t}{i})^T \boldsymbol{\alpha_i}$ as the baseline, and $f(\state{t}{i})^T \boldsymbol{\beta_i}$ as the advantage (i.e. the advantage of taking action 1 over action 0).

We re-write the reward model as follows:
\begin{align}
    \reward{t+1}{i} 
    &= \Phii{T}{it} \tparam{}{i} + \epsilon_{i}^{(t)}
\end{align}
where $\Phii{T}{it} = \Phii{}{}(\state{t}{i}, \action{t}{i}, \pii{t}{i})^T = [g(\state{t}{i})^T, (\action{t}{i} - \pii{t}{i}) f(\state{t}{i})^T, (\pi_i^{(t)})f(\state{t}{i})^T]$ is the design matrix for given state and action, and $\tparam{}{i} = [\boldsymbol{\alpha_i}, \boldsymbol{\beta_i}, \boldsymbol{\gamma_i}]^T$ is the joint weight vector that the algorithm wants to learn. We further break down the joint weight vector $\tparam{}{i}$ into two components:
\begin{align}
    \tparam{}{i} = \begin{bmatrix}
            \boldsymbol{\alpha}_i\\
            \boldsymbol{\beta}_i\\
            \boldsymbol{\gamma}_i
        \end{bmatrix}
        = \begin{bmatrix}
        \boldsymbol{\alpha}_{\text{pop}} + \boldsymbol{u}_{\alpha, i}\\
        \boldsymbol{\beta}_{\text{pop}} + \boldsymbol{u}_{\beta, i}\\
        \boldsymbol{\gamma}_{\text{pop}} + \boldsymbol{u}_{\gamma, i}
        \end{bmatrix}
        = \tpop{} + \ui{}{i}
\end{align}
Here, $\tpop{}{} = [\boldsymbol{\alpha}_{\text{pop}}, \boldsymbol{\beta}_{\text{pop}}, \boldsymbol{\gamma}_{\text{pop}}]^T$ is the population level term which is common across all the participant's reward models and follows a normal prior distribution given by $\tpop{} \sim \mathcal{N}(\muprior, \Sigprior)$. On the other hand, $\ui{}{i} = [\boldsymbol{u}_{\alpha, i}, \boldsymbol{u}_{\beta, i}, \boldsymbol{u}_{\gamma, i}]^T$ are the individual level parameters, or the \emph{random effects}, for any given participant $i$. Note that the individual level parameters are assumed to be normal by definition, i.e. $\ui{}{i} \sim \mathcal{N}(\bs{0}, \Sig{}{u})$. \Cref{sec:priors} describes how we calculate the priors and initialization values.




\subsection{Online model update procedure}
\label{rl:update}

\subsubsection{Posterior update}
\label{rl:posterior_update}

We vectorize the parameters across the $m_t$ participants who have been or are currently part of the study at time $t$, and re-write the model as:
\begin{align}
    \bs{R} &= \Phii{T}{} \bs{\theta} +\bs{\epsilon}\\
    \bs{R}_i &=
    \begin{bmatrix}
        \reward{2}{i} \\
        \reward{3}{i} \\
        \vdots\\
        \reward{t+1}{i}
    \end{bmatrix}\;\;\;
    \bs{R}=
    \begin{bmatrix}
        \bs{R_{1}} \\
        \bs{R_{2}} \\
        \vdots\\
        \bs{R_{m_t}}
    \end{bmatrix}\\
    \bs{\theta} &= \begin{bmatrix}
        \bs{\theta}_1\\
        \bs{\theta}_2\\
        \vdots\\
        \bs{\theta}_{m_t}
    \end{bmatrix} = \begin{bmatrix}
        \bs{\theta}_{\text{pop}} + \bs{u}_1\\
        \bs{\theta}_{\text{pop}} + \bs{u}_2\\
        \vdots\\
        \bs{\theta}_{\text{pop}} + \bs{u}_{m_t}
    \end{bmatrix}
    = \bs{1}_{m_t} \otimes \tpop{} + \ui{}{}\\
    \bs{\epsilon}_i &=
    \begin{bmatrix}
        \epsilon_{i}^{(1)} \\
        \epsilon_{i}^{(2)} \\
        \vdots\\
        \epsilon_{i}^{(t)}
    \end{bmatrix}\;\;\;
    \bs{\epsilon}=
    \begin{bmatrix}
        \bs{\epsilon_{1}} \\
        \bs{\epsilon_{2}} \\
        \vdots\\
        \bs{\epsilon_{m_t}}
    \end{bmatrix}\\
    \bs{u}_i &\sim \mathcal{N}(\bs{0}, \bs{\Sigma}_{u})\\
    \bs{\epsilon} &\sim \mathcal{N}(\bs{0}, \sigma_{\epsilon}^2\bs{I}_{tm_t})
\end{align}

As specified before, we assume a gaussian prior on the population level term $\tpop{} \sim \mathcal{N}(\muprior, \Sigprior)$. 
The hyper-parameters of the above model, given the definition above, are the noise variance $\sige{2}{}$ and the random effects variance $\Sig{}{u}$. 
Now, at a given decision point $t$, using  estimated values of the hyper-parameters   ($\sigma^2_{\epsilon, t}$ is the estimate of $\sige{2}{}$ and $\Sig{}{u, t}$ is the estimate of $\Sig{}{u}$), the posterior mean and covariance matrix of the parameter $\tparam{}{}$ is:
\allowdisplaybreaks
\begin{align}
    \bs{\mu^{(t)}_{\text{post}}} &= \bigg(\bs{\Tilde{\Sigma}^{-1}_{\theta, t}} + \frac{1}{\siget{2}}\bs{A} \bigg)^{-1} \bigg( \bs{\Tilde{\Sigma}^{-1}_{\theta, t}} \bs{\mu_{\theta}} + \frac{1}{\siget{2}}\bs{B} \bigg) \label{eqn:postMeanTheta} \\
    \Sig{(t)}{\text{post}} &= \bigg(\bs{\Tilde{\Sigma}^{-1}_{\theta, t}} + \frac{1}{\siget{2}} \bs{A} \bigg)^{-1} \label{eqn:postCovTheta}
\end{align}
where
\begin{align}
    \bs{A} &=
    \begin{bmatrix}
        \sum_{\tau=1}^t \Phii{}{1\tau} \Phii{T}{1\tau} & \bs{0} &\cdots &\bs{0}\\
        \bs{0} & \sum_{\tau=1}^t \Phii{}{2\tau} \Phii{T}{2\tau} & \cdots & \bs{0} \\
        \vdots & \vdots & \ddots & \vdots\\
        \bs{0} & \bs{0} & \cdots & \sum_{\tau=1}^t \Phii{}{m_t\tau} \Phii{T}{m_t\tau}
    \end{bmatrix}\\
    \bs{B} &=
    \begin{bmatrix}
        \sum_{\tau=1}^t \Phii{}{1\tau} R^{(\tau + 1)}_{1}\\
        \vdots\\
        \sum_{\tau=1}^t \Phii{}{m_t\tau} R^{(\tau + 1)}_{m_t}\\
    \end{bmatrix}\\
    \bs{\mu_{\theta}} &= 
    \begin{bmatrix}
        \bs{\muprior} \\
        \bs{\muprior} \\
        \vdots\\
        \bs{\muprior}
    \end{bmatrix} \label{eqn:mu_t0}\\
    \bs{\Tilde{\Sigma}_{\theta, t}} &= 
    \begin{bmatrix}
        \bs{\Sigprior} + \bs{\Sigma_{u,t}} & \bs{\Sigprior} & \cdots & \bs{\Sigprior}\\
        \bs{\Sigprior} & \bs{\Sigprior} + \bs{\Sigma_{u,t}} & \cdots & \bs{\Sigprior}\\
        \vdots & \vdots & \ddots & \vdots\\
        \bs{\Sigprior} & \bs{\Sigprior} & \cdots & \bs{\Sigprior} + \bs{\Sigma_{u,t}}
    \end{bmatrix} \label{eqn:sig_tt}
\end{align}

The action-selection procedure (described in section \cref{sec:act_select}) uses the Gaussian posterior distribution defined by the posterior mean $\bs{\mu^{(t)}_{\text{post}}}$ and variance $\Sig{(t)}{\text{post}}$ to determine the action selection probability $\pii{t+1}{}$ and the corresponding actions for the next time steps. 

\subsubsection{Hyper-parameter update}
\label{rl:hyperparam_update}
The hyper-parameters in the algorithm's reward model are the noise variance $\sige{2}$ and random effects variance $\Sig{}{u}$. In order to update these variance estimates at the end of decision time $t$, we use Empirical Bayes \cite{morris1983parametric} to maximize the marginal likelihood of observed rewards, marginalized over the parameters $\tparam{}{}$. So, in order to form $\Sig{}{u,t}$ and $\siget{2}$, we solve the following optimization problem:
\begin{align}
    \Sig{}{u,t}, \siget{2} &= \argmax l(\Sig{}{u,t}, \siget{2} ; \mathcal{H}^{(t)}_{1:m_t})
    \label{eqn:argmaxLL}
\end{align}
where, $\mathcal{H}^{(t)}_{1:m_t} = \{\state{1}{i}, \action{1}{i}, \reward{2}{i}, \cdots, \state{t}{i}, \action{t}{i}, \reward{t+1}{i}\}_{i \in [m_t]}$ refers to the trajectories (history of state-action-reward tuples) from time $\tau = 1$ to $\tau = t$ for all participants $i \in [m_t]$, and
\begin{align}
    l(\Sig{}{u,t}, \siget{2} ; \mathcal{H}^{(t)}_{1:m_t}) &= \log(\det(\bs{X})) - \log(\det(\bs{X} + y \bs{A})) + m_t t \log(y) - y \sum_{\tau\in [t]} \sum_{i \in [m_t]} (R^{(\tau + 1)}_{i})^2 \nonumber \\
    & - \bs{\mu_{\theta}^T} \bs{X} \bs{\mu_{\theta}} + (\bs{X} \bs{\mu_{\theta}}  + y \bs{B})^T (\bs{X} + y \bs{A} )^{-1} (\bs{X} \bs{\mu_{\theta}}  + y \bs{B})
\end{align}
Note that, $\bs{X} = \bs{\Tilde{\Sigma}^{-1}_{\theta, t}}$ (see \Cref{eqn:sig_tt}) and $y = \frac{1}{\sigma_{\epsilon, t}^2}$.






%% file: simulator.tex
\section{Action Selection Procedure}
\label{sec:act_select}
The action selection procedure utilizes a modified posterior sampling algorithm called the smooth posterior sampling algorithm. Recall from Section~\ref{rl:reward_model}, our model for the reward is a Bayesian linear mixed model with action centering given as:
\begin{equation}
    \reward{t+1}{i} = g(\state{t}{i})^T \boldsymbol{\alpha_i}  + (\action{t}{i} - \pii{t}{i}) f(\state{t}{i})^T \boldsymbol{\beta_i} + (\pi_i^{(t)})f(\state{t}{i})^T \boldsymbol{\gamma_i} + \epsilon_i^{(t)}
\end{equation}
where $\pii{t}{i}$ is the probability that the RL algorithm selects action $\action{t}{i} = 1$ in state $\state{t}{i}$ for participant $i$ at decision point $t$. The RL algorithm computes the probability $\pi_i^{(t)}$ as follows:
\begin{equation}
    \begin{aligned}
    \pi_i^{(t)} = \mathbb{E}_{{\Tilde{\beta} \sim \mathcal{N}(\mu_{\text{post}, i}^{(t-1)}, \Sigma_{\text{post}, i}^{(t-1)})}}[\rho(f(\state{t}{i})^T \boldsymbol{\Tilde{\beta}}) |\mathcal{H}_{1:m_{t-1}}^{(t-1)},  \state{t}{i}]
    \end{aligned}
\end{equation}
where $\mathcal{H}^{(t-1)}_{1:m_{t-1}} = \{\state{1}{i}, \action{1}{i}, \reward{2}{i}, \cdots, \state{t-1}{i}, \action{t-1}{i}, \reward{t}{i}\}_{i \in [m_{t-1}]}$ refers to the trajectories (history of state action reward tuples) from time $\tau = 1$ to $\tau = t-1$ for all participants $i \in [m_{t-1}]$.
Notice that the last expectation above is only over the draw of $\beta$ from the posterior distribution parameterized by $\boldsymbol{\mu_{\text{post}, i}^{(t-1)}}$ and $\boldsymbol{\Sigma_{\text{post}, i}^{(t-1)}}$ (see \cref{eqn:postMeanTheta} and \cref{eqn:postCovTheta} for their definitions).

\begin{figure}[!h]
    \centering
    \begin{subfigure}[t]{0.49\textwidth}
        \centering
        \includegraphics[width=0.7\textwidth]{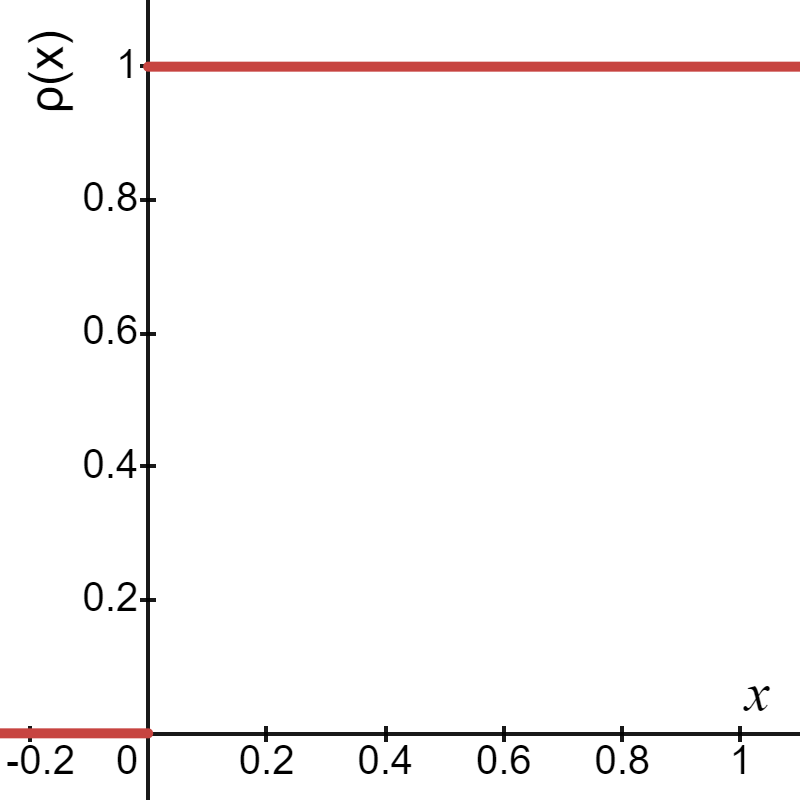}
        \caption{Indicator function}
        \label{fig:indicator_func}
    \end{subfigure}
    \hfill
    \centering
    \begin{subfigure}[t]{0.49\textwidth}
        \centering
        \includegraphics[width=0.7\textwidth]{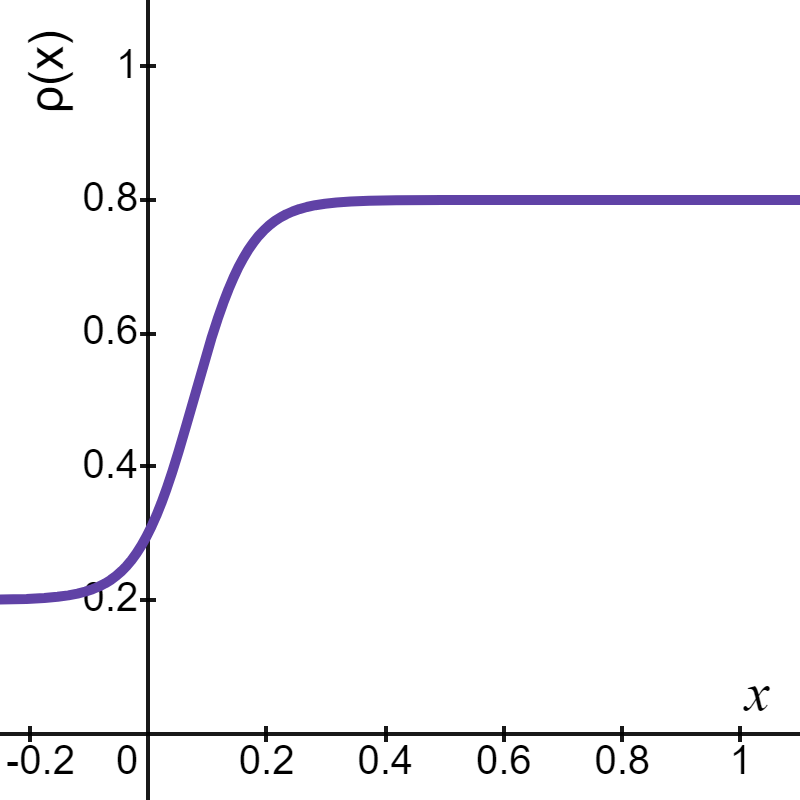}
        \caption{Generalized logistic function}
        \label{fig:smooth_alloc_func}
    \end{subfigure}
    \caption{Posterior sampling allocation functions}
    \label{fig:alloc_funcs}
\end{figure}

In classical posterior sampling, the posterior sampling algorithm uses an indicator function (\Cref{fig:indicator_func}):
\begin{equation}
    \rho(x) = \II(x > 0)
    \label{eqn:classical_post_sampling1}
\end{equation}
If the indicator function above is used, the posterior sampling algorithm sets randomization probabilities to the posterior probability that the treatment effect is positive. However, in order to facilitate after-study analysis while using a pooled algorithm, past works have shown that it is better to replace the indicator function (Equation \ref{eqn:classical_post_sampling1}) with a \emph{smooth} i.e. continuously differentiable function \cite{zhang2022statistical}. 
Using a smooth function $\rho$ ensures the policies formed by the algorithm concentrate. Concentration enhances the replicability of the randomization probabilities if the study is repeated. 
Without concentration, the randomization probabilities might fluctuate greatly between repetitions of the same study. For more discussion of this see \cite{deshpande2018accurate,zhang2020inference,NEURIPS2021_49ef08ad,zhang2022statistical}. In MiWaves, we choose $\rho$ to be a generalized logistic function (\Cref{fig:smooth_alloc_func}), defined as follows:
\begin{equation}
    \rho(x) = L_{\min} + \frac{ L_{\max} - L_{\min} }{  1 + c \exp(-b x) }
    \label{eqn:smooth_post_sampling0}
\end{equation}
where
\begin{itemize} 
    \item where $L_{\min} = 0.2$ and $L_{\max} = 0.8$ are the lower and upper clipping probabilities  (i.e., $0.2\le\pi_i^{(t)}\le 0.8$).
    \item $c=5$. Larger values of $c > 0$ shifts the value of $\rho(0)$ to the right.  This choice implies that $\rho(0)=0.3$. Intuitively, the probability of taking an action when the current estimate of the treatment effect is approximately $0$, is $0.3$.
    \item $b = 21.053$. Larger values of $b > 0$ makes the slope of the curve more ``steep''. Please refer to \cref{sec:smooth_allocation_function} for more details as to how we arrive at this value.
\end{itemize}

\section{Prior data and MiWaves Simulation Testbed}
\label{appendix:testbed}
This section details how we transform prior data to construct a dataset, and utilize the dataset to develop the MiWaves simulation testbed. The testbed is used to develop and evaluate the design of the RL algorithm for the MiWaves study. The \emph{base simulator} or the \emph{vanilla testbed} is constructed using the SARA \cite{rabbi2018toward} study dataset. The SARA dataset consists of $N=70$ participants, and the SARA study was for 30 days, 1 decision point per day. For each participant, the dataset contains their daily and weekly survey responses about substance use, along with daily app interactivity logs and notification logs. We will now detail the procedure to 
construct this base simulator.

\subsection{SARA vs MiWaves}
The Substance Use Research Assistant (SARA) \cite{rabbi2018toward} study trialed an mHealth app aimed at sustaining engagement of substance use data collection from participants. Since the SARA study focused on a similar demographic of emerging adults (ages 18-25) as the MiWaves study, we utilized the data gathered from the SARA study to construct the simulation testbed. We highlight the key differences between the SARA study and the MiWaves study in \Cref{tab:sara_vs_miwaves}.

\begin{table}[!h]
    \centering
    \begin{tabular}{|P{0.45\textwidth}|P{0.45\textwidth}|}
        \hline
        SARA & MiWaves \\
        \hline
        \hline
        \bo{70} participants & \bo{120} participants planned\\
        \hline
        \bo{One} decision point per day & \bo{Two} decision points per day\\
        \hline
        Cannabis use data collected \bo{weekly}	& Cannabis use data collected \bo{daily} \\
        \hline
        Micro-randomization: \bo{0.5} probability & 	Micro-randomization: \bo{Determined by RL} \\
        \hline
        Action: \bo{Inspirational messages only} & Action: Messages of \bo{varying prompt length and interaction levels to prompt reduction of cannabis use} \\
        \hline
        Goal of intervention messages: \bo{to increase survey completion to collect substance use data} & Goal of intervention messages: \bo{to reduce cannabis use through self-monitoring and mobile health engagement}\\
        \hline
    \end{tabular}
    \caption{SARA vs MiWaves: Key differences}
    \label{tab:sara_vs_miwaves}
\end{table}

\subsection{Data Extraction}
\label{sec:data_extract}
First, we will detail the steps to extract the relevant data from the SARA dataset:
\begin{enumerate}
    \item  \bo{App Usage:} The SARA dataset has a detailed log of the participant's app activity for each day in the trial, since their first login. We calculate the time spent by each participant between a normal entry and an \emph{app paused} log entry, until 12 AM midnight, to determine the amount of time (in seconds) spent by the participant in the app on a given day. To determine the time spent by the participant in the evening, we follow the same procedure but we start from any logged activity after the 4 PM timestamp, till midnight (12 AM). We discuss the procedure to compute time spent by the participant in the morning in \Cref{sec:dataset_gen}. 
    \item \bo{Survey Completion:} The SARA dataset contains a CSV file for each participant detailing their daily survey completion status (completed or not). We use this binary information directly to construct the survey completion feature.
    \item \bo{Action:} The SARA dataset contains a CSV file for each participant detailing whether they got randomized (with 0.5 probability) to receive a notification at 4 PM, and whether the notification was actually pushed and displayed on the participant's device. We use this CSV file to derive the latter information, i.e. whether the app showed the notification on the participant's device (yes or no).
    \item \bo{Cannabis Use:} Unlike MiWaves, the SARA trial did not ask participants to self-report cannabis use through the daily surveys. However, the participants were prompted to respond to a weekly survey at the end of each week (on Sundays). Through these weekly surveys, the participants were asked to retroactively report on their cannabis use in the last week, from Monday through Sunday. We use these weekly surveys to retroactively build the reported cannabis-use for each participant in the study. The cannabis use was reported in \emph{grams} of cannabis used, taking values of $0$g, $0.25$g, $0.5$g, $1$g, $1.5$g, $2$g and $2.5$g+. Participants who were not sure about their use on a particular day, reported their use as ``\emph{Not sure}''. Meanwhile, participants who did not respond to the weekly survey were classified to have ``\emph{Unknown}'' use for the entire week. The distribution of reported cannabis use in SARA across all the participants, across all 30 days, can be viewed in Figure \ref{fig:cb_use_dist}.

    \begin{figure}[h]
        \includegraphics[scale=0.5]{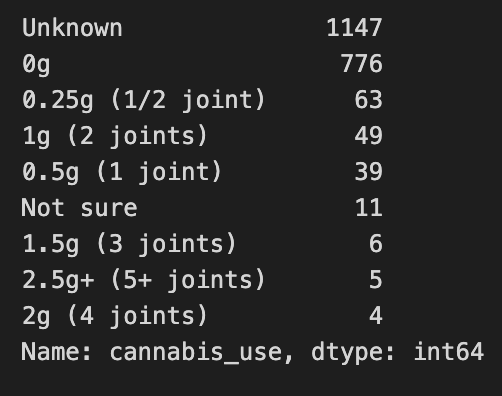}
        \centering
        \caption{Distribution of cannabis use across all participants, across all data points}
        \label{fig:cb_use_dist}
    \end{figure}
\end{enumerate}

We build a database for all participants using the three features specified above.

\subsection{Data Cleaning}
Next, we will specify the steps to clean this data, and deal with outliers:

\begin{enumerate}
    \item \bo{Participants with insufficient data}: We remove participants who had more than 20 undetermined (i.e. either ``\emph{Unknown}'' or ``\emph{Not sure}'') cannabis use entries. Upon removing such participants, we are left with $N=42$ participants. The updated distribution of reported cannabis use in the remaining data across all the participants across all 30 days, is demonstrated in Figure \ref{fig:cb_use_dist_filter}.

    \begin{figure}[!h]
        \centering
        \begin{subfigure}[t]{0.49\textwidth}
            \centering
            \includegraphics[width=0.51\textwidth]{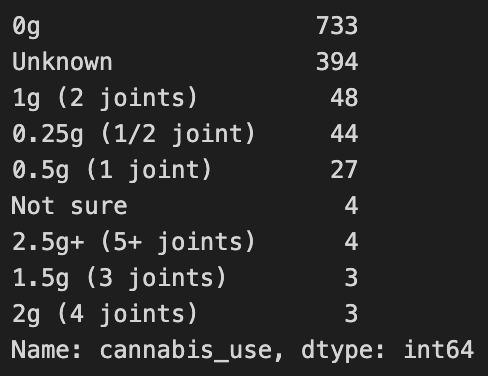}
            \caption{Dataset cannabis-use distribution}
            \label{fig:dataset_cb_dist}
        \end{subfigure}
        \hfill
        \centering
        \begin{subfigure}[t]{0.49\textwidth}
            \centering
            \includegraphics[width=\textwidth]{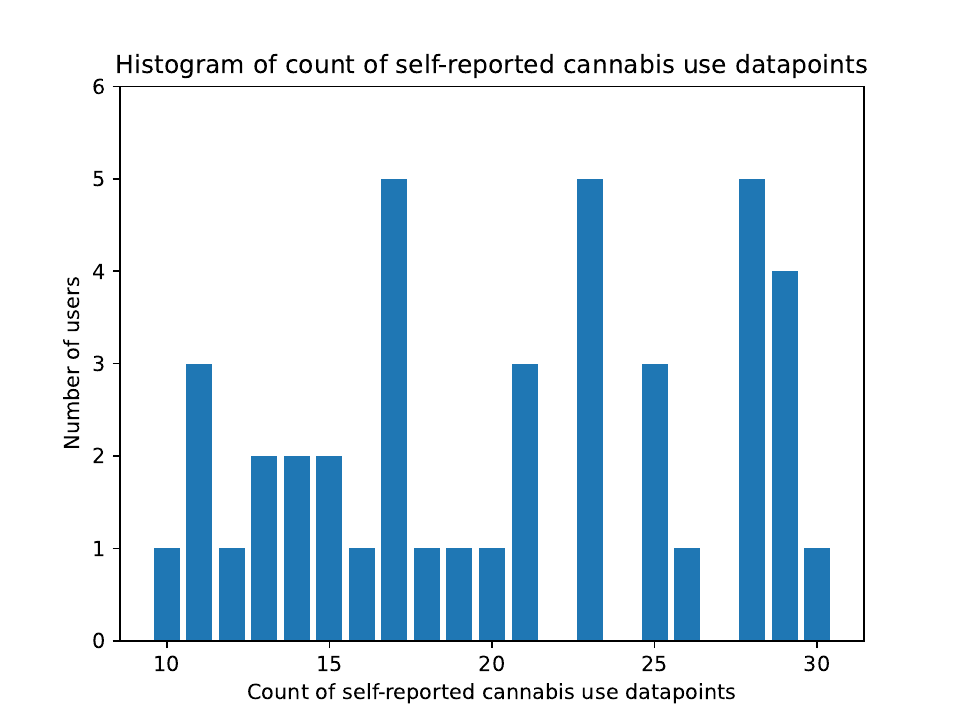}
            \caption{Count of non-missing cannabis-use datapoints vs number of participants}
            \label{fig:non-missing_count}
        \end{subfigure}
        \caption{Distribution of cannabis use across all participants, across all data points, after removing participants with insufficient data}
        \label{fig:cb_use_dist_filter}
    \end{figure}

    \item \bo{Outliers in app usage information}: The average app usage in a given day across all participants, comes out to be $244$ seconds (please refer to Section \ref{sec:data_extract} as to how app usage is calculated from the raw data).
    However, sometimes, a participant ended up with greater than $1000$ seconds of app use in a day, with the highest reaching around $67000$ seconds. We observe a similar issue with the evening app use data. To deal with such outliers, we clip any evening app use higher than $700$ seconds. There are $20$ such evening app use data points in the dataset.
\end{enumerate}

\subsection{Reward computation}
\label{sec:reward_comp}
Next, we calculate the \emph{reward} for each data point in the dataset. It is calculated as follows:
\begin{itemize}
    \item \bo{0}: Participant did not complete the daily survey, nor used the app.
    \item \bo{1}: Participant did not complete the daily survey, but used the app outside of the daily survey.
    \item \bo{2}: Participant completed the survey.
\end{itemize}

We transform the $[0-2]$ reward above to a $[0-3]$ reward defined in MiWaves (refer \cref{sec:rl_framework}) by randomly transforming the data points with $2$ reward to $3$ with $50\%$ probability.

\subsection{Dataset for training participant models}
\label{sec:evening_dataset}
We use the post-4PM data (i.e. data from 4 PM to 12 AM midnight) 
(dubbed as the \emph{evening} data) to create a dataset to train the individual participant models for the MiWaves simulation testbed. The dataset contains the following features:
\begin{itemize}
    \item Day In Study $\in\;[1, 30]$
    \item Cannabis Use $\in\;[0, 2.5g]$
    \item (Evening) App usage $\in\;[0, 700s]$
    \item Survey completion (binary)
    \item Weekend indicator (binary)
    \item Action (binary)
    \item Reward $\in \{0, 1, 2, 3\}$
\end{itemize}
We detail the steps to create this \emph{evening} dataset:

\begin{enumerate}
    \item \bo{Evening cannabis-use}: The SARA study documented cannabis use for a given participant for a whole day. In contrast, MiWaves will be asking participants to self-report their cannabis use twice a day. To mimic the same, after discussions among the scientific team,
    we split a given participant's daily cannabis use from SARA into morning and evening use.  In particular we multiply the total day's use by a factor of $0.67$ to generate their evening cannabis use. Also, the MiWaves study will be recruiting participants who typically use cannabis at least 3 times a week. We expect their use to be much higher than that of participants in SARA. So, we multiply the evening cannabis use again by a factor of $1.5$. Thus, we generate the evening cannabis use from the participant's daily use reported in SARA as follows:
    \begin{equation}
        \text{Evening CB Use} = \text{That Day's SARA CB Use} \times 0.67 \times 1.5
    \end{equation}
    
    \item \bo{Feature normalization}: The resulting dataset's features are then normalized as follows:
    \begin{itemize}
        \item \bo{Day in study} is normalized into a range of $[-1, 1]$ as follows
        \begin{align}
            \text{Day in study (normalized)} = \frac{\text{Day in study} - 15.5}{14.5}
        \end{align}
        \item \bo{App usage} is normalized into a range of $[-1, 1]$ as follows
        \begin{align}
            \text{App usage (normalized)} = \frac{\text{App usage} - 350}{350}
        \end{align}
        \item \bo{Cannabis use} (evening) is normalized into a range of $[-1,1]$ as follows
        \begin{align}
            \text{Cannabis use (normalized)} = \frac{\text{Cannabis use} - 1.3}{1.35}
        \end{align}
    \end{itemize}
\end{enumerate}

\subsection{Training Participant Models}
\label{sec:training_user_models}

As specified above in Sec. \ref{sec:evening_dataset}, we use the \emph{evening} dataset to train our participant models for the MiWaves simulation testbed. This dataset has the following features:  
\begin{itemize}
    \item Day In Study $\in\;[-1, 1]$: Negative values refer to the first half of the study, while positive values refer to the second half of the study. A value of $0$ means that the participant is in the middle of the study. $-1$ means that the participant has just begun the study, while $1$ means they are at the end of the 30 day study. 
    \item Cannabis Use $\in\;[-1, 1]$: Negative values refer to the participant's cannabis use being lower than the population's average cannabis use value, while positive values refer to participant's cannabis use being higher than the study population's average cannabis use value. A value of $0$ means that the participant's cannabis use is the average value of cannabis use in the study population. Meanwhile, $-1$ means that the participant is not using cannabis, and $1$ means that the participant used the highest amount of cannabis reported by the study population. 
    \item (Evening) App usage $\in\;[-1, 1]$: Negative values refer to the participant's app use being lower than the population's average app use value, while positive values refer to participant's app use being higher than the study population's average app use value. A value of $0$ means that the participant's app usage is the average amount of app usage observed in the study population. Meanwhile, $-1$ means that the participant's app usage is non-existent (i.e. zero). On the other hand, $1$ means that the participant's app usage is the highest among the observed app usage values in the study population. 
    \item Survey completion $\in \{0, 1\}$: A value of $0$ refers to the case where the participant has not finished the decision point's check-in, while a value of $1$ refers to the case where the participant has responded to the decision point's check-in.
    \item Weekend indicator $\in \{0, 1\}$: A value of $0$ refers to the case where the decision point falls on a weekday, while a value of $1$ refers to the case where the decision point falls on a weekend.
    \item Action $\in \{0, 1\}$: A value of $0$ means that action was not taken (i.e. no notification or intervention message was shown), while $1$ means that an action was taken (i.e. a notification or intervention message was shown).
    \item Reward $\in \{0, 1, 2, 3\}$: Same as defined in Section \ref{sec:reward_comp}
\end{itemize}

We fit the reward using our participant models. Before we train participant models, we do a \emph{complete-case analysis} on the \emph{evening} dataset. It involves removing all the participant-evening data points which have \bo{any} missing feature. This can either be a missing ``cannabis use'' value, or a missing ``action'' (which is the case on the first day of the study). To that end, 435 participant-evening data points are removed out of a total of 1260 participant-evening data points.
Given that our target variable i.e. the reward is categorical (0-3) in nature, we consider two options for our generative participant models:
\begin{itemize}
    \item \bo{Multinomial Logistic Regression (MLR)} - We fit a multinomial logistic regression model on our dataset. Given $K$ classes (K=4 in our case), the probability for a data point $i$ belonging to a class $c$ is given by:

    \begin{align}
        P(Y_i = c|X_i) = \frac{e^{\beta_c \cdot X_i}}{\sum_{j=1}^{K} e^{\beta_j \cdot X_i}}
    \end{align}
    where $X_i$ are the features of the given \emph{data point} $i$, and $\beta_j$ are the learned weights for a given reward class $j$. $\dim(\beta_j) = \dim(X_i) = M$, i.e. each data point has a dimension of $M$ features. Each \emph{data point} refers to a participant-decision point, we use these terms interchangibly throughout the document.
    
    The model is optimized using the following objective \cite{1-Linear}:
    \begin{align}
        \min_{\beta} - \sum_{i=1}^{n} \sum_{c=1}^{K} \mathbbm{1}_{\{Y_i = c\}} \log (P(Y_i = c|X_i)) + \frac{1}{2} ||\beta||^2_{F}
    \end{align}

    where,
    \begin{align}
        ||\beta||^2_{F} = \sum_{i = 1}^{M} \sum_{j=1}^{K} \beta_{i,j}^2
    \end{align}
    We use python's \texttt{scikit-learn} package for training the multinomial logistic regression model (more information \href{https://scikit-learn.org/stable/modules/linear_model.html#multinomial-case}{here}). It makes sure that $\sum_{j} \beta_j = 0$. We use the following parameters for training the model using \texttt{scikit-learn}: 
    \begin{itemize}
        \item \bo{Penalty:} L2
        \item \bo{Solver:} LBFGS
        \item \bo{Max. iterations:} 200
        \item \bo{Multi-class:} Multionomial
    \end{itemize}
    \item \bo{Multi-layer perceptron (MLP) Classifier} - We fit a simple neural network on our dataset, and use the last layer's logits as probabilities for each class.
    
    We use python's \texttt{scikit-learn} package for training the MLP Classifier. We use the following configuration while training the MLP classifier: 
    \begin{itemize}
        \item \bo{Hidden layer configuration:} (7, )
        \item \bo{Activation function:} Logistic
        \item \bo{Solver:} Adam
        \item \bo{Max. iterations:} 500
    \end{itemize}
\end{itemize}

\begin{figure}[t]
    \centering
    \includegraphics[width=0.99\textwidth]{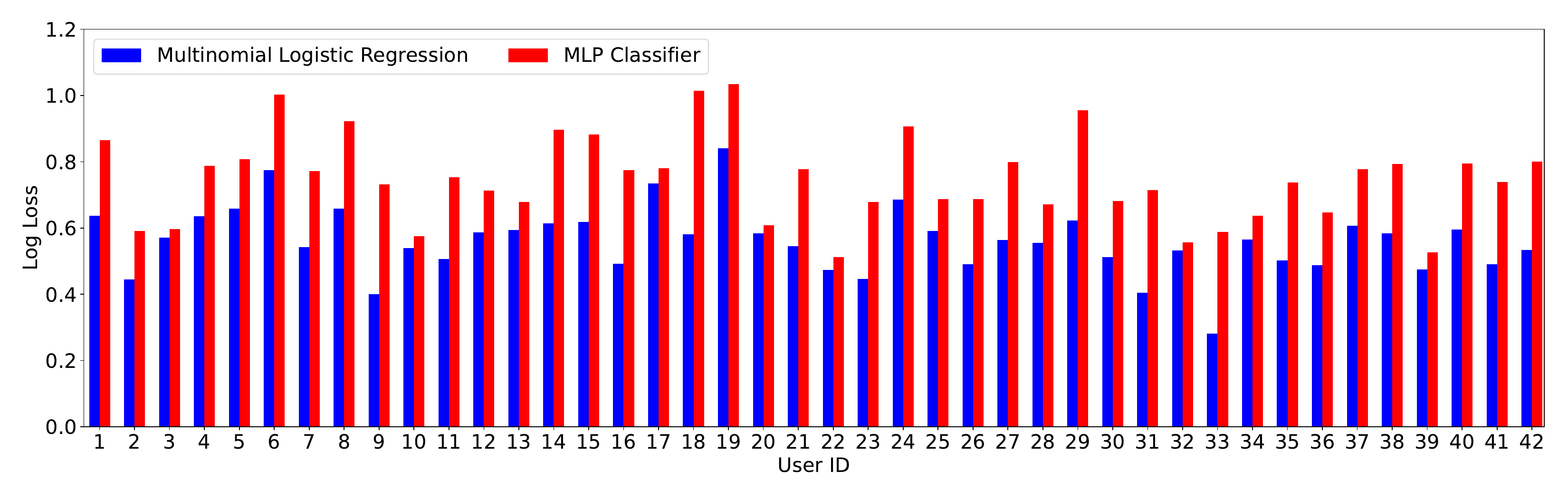}
    \caption{Comparison of log loss between the two models across all participants}
    \label{fig:log_loss}
\end{figure}

We choose MLR for our participant models, as it is interpretable, and offers similar (if not better) performance as compared to a generic neural network (see Figure \ref{fig:log_loss}). Interpretability is important here since we would like to construct testbed variants with varying treatment effects.

\begin{landscape}
\begin{figure}[ht]
    \centering
    \includegraphics[width=1.2\textwidth]{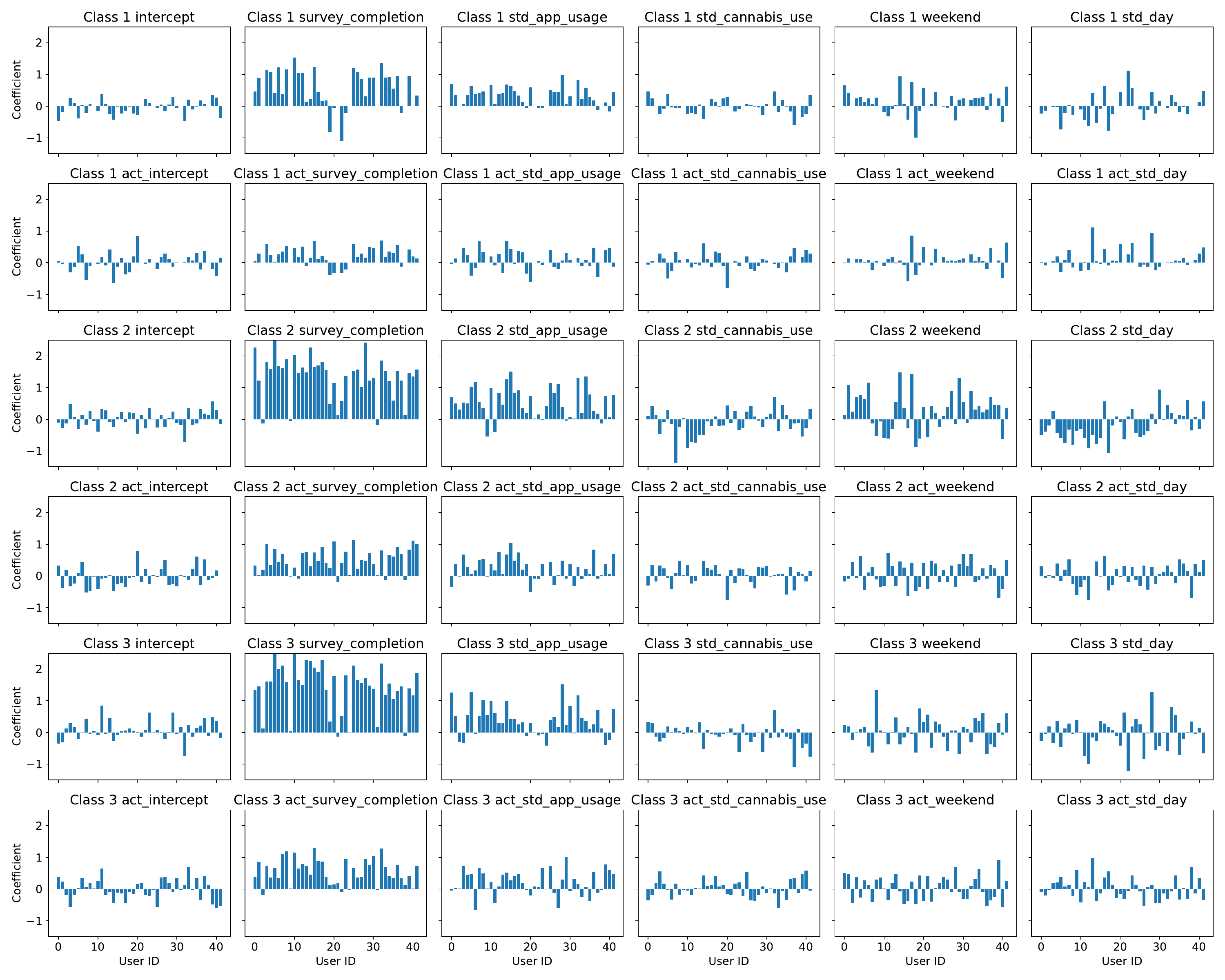}
    \caption{\textbf{Bar plot of coefficients of features in the MLR participant models relative to coefficients of class 0, across all $N=42$ participants.}}
    \label{fig:coeff_relative}
\end{figure}

\end{landscape}

Figure \ref{fig:coeff_relative} represents the learnt coefficients of the MLR participant model for classes 1 to 3, relative to class 0's coefficients. Note that both \emph{survey completion} and \emph{app usage} seem to exhibit strong relationship wrt the target variable for most of the participants. To be specific, in Figure \ref{fig:coeff_relative}, coefficients of both \emph{survey completion} and \emph{app usage} are mostly positive across most of the $N=42$ participants, both in baseline and advantage. The magnitude of the relative weights of these features keeps increasing as the reward class increases. This signifies that if a participant is engaged (completing surveys, using the app), they are more likely to generate a non-zero reward as compared to a reward of 0. We use the probabilities as weights from the MLR models to stochastically generate the reward during the data generation process. More on that in Section \ref{sec:data_gen}.

\newgeometry{margin=0.1cm} 
\restoregeometry

\subsection{Dataset for generative process}
\label{sec:dataset_gen}
Next, we will create a dataset for the generative process for a simulation. To that end, we impute missing values in the \emph{evening} dataset, and also create the \emph{morning} dataset. We describe the procedure for both below.

\begin{itemize}
    \item \bo{Imputation in \emph{evening} dataset}: As stated before, the vanilla \emph{evening} dataset has a few missing values for ``cannabis use'' and ``action'' values. We impute the values for ``cannabis use'' as follows - for a given missing value, we first determine the day of the week, and then replace the unknown/missing value with the average of the cannabis use across the available data of the participant for that day of the week.
    
    Note that during the simulation, the action used by the participant models for reward generation will be determined by the RL algorithm in the simulation. Hence, we do not need to impute ``action'' here.

    \item \bo{Generating \emph{morning} dataset}: Similar to the \emph{evening} dataset, we generate a \emph{morning} dataset per participant to mimic the data we would receive from MiWaves (2 decision points a day). We generate the following features:
    \begin{itemize}
        \item \bo{Cannabis Use}: We generate the morning cannabis use as follows:
        
        \begin{equation}
            \text{Morning CB Use} = \text{That Day's SARA CB Use} \times 0.33 \times 1.5
        \end{equation}
        \item \bo{App Usage}: Since there are less than 30 evening app usage values per participant in the SARA dataset, we decide to use these values as an empirical distribution, and resample, with replacement, from the 30 values for each participant at each morning. The sampled value is the participant's  morning app usage.

        \item \bo{Survey Completion}: For each participant, we determine the proportion of times they responded to the daily survey during the SARA study. Using this ratio as our probability of survey completion, we sample from a binomial distribution to construct the Survey Completion feature for the morning dataset for each participant for each day.

        \item \bo{Day In Study and Weekend indicator}: We create one morning data point per day, so the day in study and weekend indicator features mimic the evening dataset. 
    \end{itemize}
\end{itemize}

We combine both the \emph{morning} and \emph{evening} dataset, and we call it the \emph{combined} dataset. We will now describe the procedure to generate data during simulations using this dataset.

\subsection{Simulations: Data generation}
\label{sec:data_gen}
In this section, we detail our data generation process for a simulation. We assume that any given participant $i$, her trajectory in an RL simulation environment with $T$ total decision points has the following structure: $\mathcal{H}_{i}^{(T)} = \{\state{1}{i}, \action{1}{i}, \reward{2}{i}, \cdots, \state{T}{i}, \action{T}{i}, \reward{T+1}{i}\}$.

We also assume that the \emph{combined} dataset has the following structure - for any given decision point $t$ ($t \in [1,60])$, the state, $\state{t}{i}$, is constructed (partially) based on: the participant's \emph{app usage} from $t-1$ to $t$, the \emph{survey completion} indicator (whether participant fills the survey after receiving action at $t$) for decision point $t$, and the \emph{cannabis use} of the participant from $t-1$ to $t$. The data at decision point $t$ from the \emph{combined} dataset is used to generate $\reward{t+1}{i}$, which in turn helps generate features to form $\state{t+1}{i}$ (refer Section \ref{sec:rl_framework}).

\begin{enumerate}
    \item Given a set number of participants (parameter of the simulator) to simulate, we sample participants with replacement from the $N=42$ participants in the \emph{combined} dataset.
    \item We start the simulation in the morning of a given day. From decision point $t=1$ to $T$ in the simulation ($T=60$), for each sampled participant $i$ (refer to the RL framework in Section \ref{sec:rl_framework}):
    \begin{enumerate}
        \item If $t>1$, given previously-generated reward $\reward{t}{i}$, we construct the following features - \emph{survey completion}, \emph{app usage indicator}, and \emph{activity question} (which will help form $\state{t}{i}$ according to Section \ref{sec:rl_framework}), according to the following rules:
        \begin{align}
            \text{Survey Completion} = 
            \left\{
        	\begin{array}{ll}
        		1  & \mbox{if } \reward{t}{i} \geq 2 \\
        		0 & \mbox{if } \reward{t}{i} < 2
        	\end{array}
            \right.
            \label{eqn:survey_comp}
        \end{align}
        \begin{align}
            \text{App Usage Indicator} = 
            \left\{
        	\begin{array}{ll}
        		1  & \mbox{if } \reward{t}{i} \geq 1 \\
        		0 & \mbox{if } \reward{t}{i} = 0
        	\end{array}
            \right.
            \label{eqn:app_use}
        \end{align}
        \begin{align}
            \text{Activity Question} = 
            \left\{
        	\begin{array}{ll}
        		1  & \mbox{if } \reward{t}{i} = 3 \\
        		0 & \mbox{otherwise}
        	\end{array}
            \right.
            \label{eqn:act_ques}
        \end{align}
        
        We communicate the aforementioned calculated features (Equation \ref{eqn:survey_comp}, \ref{eqn:app_use}, and \ref{eqn:act_ques}), along with the \emph{cannabis use} from $t-2$ to $t-1$ to the RL algorithm. This is similar to how a participant would self-report and convey this information to the app, to help the algorithm derive the participant current state $\state{t}{i}$.

        At $t=1$, since there is no $\reward{1}{i}$, note that the RL algorithm uses some initial state $\state{1}{i}$. Since we intend to simulate MiWaves, all RL algorithms will use the following $\state{1}{i}$:
        \begin{itemize}
            \item $S_{i, 1}^{(1)}$: Set to 0. At the start of the trial, there is no engagement by the participant.
            \item $S_{i, 2}^{(1)}$: Set to 0. The first decision point is the morning, which corresponds to 0.
            \item $S_{i, 3}^{(1)}$: Set to 1. The participants in MiWaves use cannabis regularly (at least 3x a week), so we ascertain the participants in the trial to have high cannabis use at the start of the trial.
        \end{itemize}
        
        \item We ask the RL algorithm for the action $\action{t}{i}$ to be taken at the current decision point $t$ for participant $i$.
        \item We take the data from the $t^{th}$ decision point in \emph{combined} dataset of the participant $i$ - specifically the participant's \emph{survey completion} at $t$, the participant's \emph{app usage} from $t-1$ to $t$, and the participant's \emph{cannabis use} from the $t-1$ to $t$. We calculate the \emph{weekend indicator} by checking whether the decision point $t$ falls on Saturday or Sunday. We feed this data, along with the action $\action{t}{i}$ from the previous step, to the participant $i$'s trained participant model from Section \ref{sec:training_user_models} to obtain a reward $\reward{t+1}{i}$.
        
    \end{enumerate}
\end{enumerate}

\subsection{Environment Variants}
\label{sec:env_variants}
We will now enumerate the different environment variants we have planned for our simulator, and how we go about operationalizing them. Since we know that the MiWaves study is planned to recruit around 120 participants, all our simulation environments will have $120$ participants. We sample these participants with replacement from the $N=42$ participants from the \emph{combined dataset} during the data generation process.

\begin{enumerate}
    \item \bo{Varying size of the treatment effect:} We construct three variants wrt the size of the treatment effect:
    \begin{itemize}
        \item \bo{Minimal treatment effect size}: We keep the treatment effect i.e. advantage coefficients in the MLR model that we learn from the participants in the SARA data.
    \end{itemize}
    
    For the other two variants, we modify the advantage weights for each of the MLR participant models. Specifically, we only modify the advantage intercept weights. We find the minimum advantage intercept weight across all classes learnt using the SARA participant data, and if it is not assigned to class 0 - we swap the weight with that of class 0. Then we set the advantage intercept weight of class 2 and class 3 to be the average of both.
    The reason we do so, is the fact that we believe that taking an action will always have a non-negative treatment effect. To that end, taking an action will always result in higher probabilities for generating a non-zero reward, and lower probabilities for generating a zero reward. Setting class weights by assigning the minimum weight to reward class 0 helps us achieve that. Also, since we trained the participant models from SARA data where we assigned 2 and 3 reward with equal probability wherever reward of 2 was assigned, we set their corresponding advantage intercept weights to be equal. Hence, we set them to be the average of the observed weights of the two reward classes. Moreover, we decide to work with the participant's learnt weights from SARA to preserve heterogeneity among participants. 

    We then multiply the advantage intercept weights for all the classes by a factor, which we refer to as the \emph{multiplier}. We select the multipliers for our other variants after observing the \emph{standardized effect sizes}.
    To compute the standardized effect size, we first generate a dataset of the $N=42$ participants' trajectories (from SARA) for each multiplier, by sampling actions with probability $0.5$. We generate 500 datasets (i.e. 500 simulations with $N=42$ participants in each simulation) for each multiplier based environment. For each dataset (i.e each simulated trial) of $N=42$ participants, we fit a linear model with fixed effects and robust standard errors. We utilize a linear model with fixed effects to compute the standardized effect size since the RL algorithm's reward approximation model (with random effects) has the same form in expectation. 
    The model is specified as follows:
    \begin{align}
        \notag E\left[\reward{t+1}{i}|S_1, S_2, S_3, a\right] &= \alpha_0 + \alpha_1 S_1 + \alpha_2 S_2 + \alpha_3 S_3 + \alpha_4 S_1 S_2 + \alpha_5 S_1 S_3 + \alpha_6 S_2 S_3 + \alpha_7 S_1 S_2 S_3\\
        &+ a  (\beta_0 + \beta_1 S_1 + \beta_2 S_2 + \beta_3 S_3 + \beta_4 S_1 S_2 + \beta_5 S_1 S_3 + \beta_6 S_2 S_3 + \beta_7 S_1 S_2 S_3).
        \label{eqn:sanity_check_reward_model}
    \end{align}
    
    \begin{figure}[!ht]
        \centering
        \includegraphics[width=0.45\textwidth]{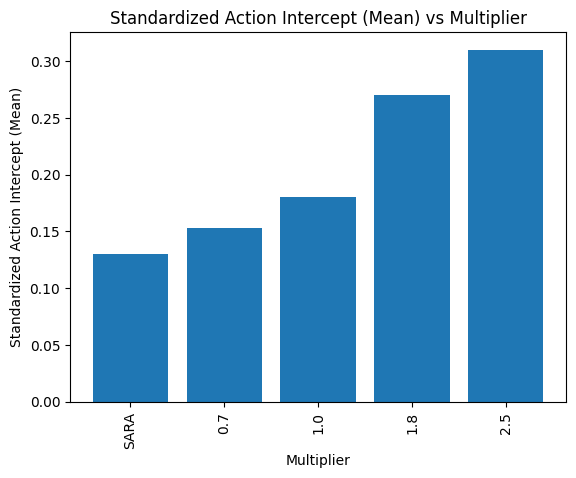}
        \caption{Comparison of mean standardized effect size vs advantage intercept multiplier. Mean is taken across 500 simulated trials each with $N=42$ participants (from SARA) in each simulation}
        \label{fig:standardized_act_intercept}
    \end{figure}
    

    To obtain the \emph{standardized effect size}, we take the average value of the advantage terms and divide it by the sample standard deviation of observed rewards (i.e. for that simulated trial) of $N=42$ participants. We do so for all the different multipliers we consider, and the results are summarized in Fig \ref{fig:standardized_act_intercept}. The standardized action intercept represents the increase in the average reward when a notification is delivered with $S_1 = S_2 = S_3 = 0$. We first check the minimal effect (weights from SARA), and see that the standardized treatment effect for the minimum effect setting comes out to be around $0.12$. We want our low treatment effect setting to be closer to $0.15$ and our higher to be around $0.3$. So we tinker our weights as mentioned in the procedure above, and re-run the simulations. Since we do not scale the weights (yet) by any factor, this is equivalent to a multiplier setting of $1.0$. We observe a standardized treatment effect of $0.18$ in this case. Using this as a reference point, we try a lower and higher multiplier, and we observe that $0.7$ and $2.5$ give us the desired standardized treatment effect for low effect and high effect settings respectively.
    
    Now, the remaining variants are described as follows:
    \begin{itemize}
        \item \bo{Low}:  We multiply the advantage coefficients for each of the MLR participant models for each class by $0.7$.
        
        \item \bo{High}: We multiply the advantage coefficients for each of the MLR participant models (as mentioned above) for each class by $2.5$. This way, we are increasing the size of all advantage intercepts in the MLR participant model, and in turn further skews the probability of getting a higher reward when taking an action as compared to a zero reward.
    \end{itemize}

    


    \item \bo{Differential treatment effect in morning vs evening}: We have two variants wrt. this:
    \begin{itemize}
        \item Set the treatment effect size of mornings to be Low, and evenings to be High (Low and High wrt Variant 4 described above in ``Varying size of the treatment effect'').
        \item Set the treatment effect size of mornings to be high and evenings to be Low.
    \end{itemize}

    \item \bo{Decay in treatment effect across time}: We linearly reduce the advantage intercept weight multiplier, so that it is set to $0$ by the last day of the simulated trial.
\end{enumerate}

We design a total of 9 environment variants by combining the aforementioned environments with each other:
\begin{itemize}
    \item Minimal treatment effect (learnt using SARA data)
    \item Low treatment effect
    \item High treatment effect
    \item Low treatment effect in the morning, and high treatment effect in the evening
    \item High treatment effect in the morning, and low treatment effect in the evening
    \item Low treatment effect, with treatment effect decreasing linearly to 0 by the end of the simulated trial
    \item High treatment effect, with treatment effect decreasing linearly to 0 by the end of the simulated trial
    \item Low treatment effect in the morning, and high treatment effect in the evening, with treatment effect decreasing linearly to 0 by the end of the simulated trial
    \item High treatment effect in the morning, and low treatment effect in the evening, with treatment effect decreasing linearly to 0 by the end of the simulated trial
\end{itemize}

%% file: rl_design.tex
\section{RL Algorithm Design}
\label{chap:rl_alg_design}

This section details the RL algorithm design decisions with respect to the MiWaves study. These involve determining the priors (\cref{sec:priors}), empirically verifying the prior values (\cref{sec:empirical_check}), designing algorithm variants (\cref{sec:alg_variants}) to benchmark and finalize hyper-parameters, and describing the RL algorithm's action selection strategy (\cref{sec:smooth_allocation_function}).
The algorithm variants are benchmarked using the MiWaves simulation testbed (described in \cref{appendix:testbed}). The results along with the final algorithm are described in \cref{sec:results}.


\subsection{Initial values and Priors}
\label{sec:priors}
This section details how we calculate the initial values and priors using SARA data (see \cref{appendix:testbed}) to estimate the reward, given the state. First, we define the model for the conditional mean of the  reward $\reward{t+1}{i}$ of a given participant $i$ at time $t$, given the current state $S = \{S_1, S_2, S_3\}$ (dropping the participant index and time for brevity):
\begin{align}
    \notag E\left[\reward{t+1}{i}|S_1, S_2, S_3, a\right] &= \alpha_0 + \alpha_1 S_1 + \alpha_2 S_2 + \alpha_3 S_3 + \alpha_4 S_1 S_2 + \alpha_5 S_1 S_3 + \alpha_6 S_2 S_3 + \alpha_7 S_1 S_2 S_3\\
    &+ a  (\beta_0 + \beta_1 S_1 + \beta_2 S_2 + \beta_3 S_3 + \beta_4 S_1 S_2 + \beta_5 S_1 S_3 + \beta_6 S_2 S_3 + \beta_7 S_1 S_2 S_3).
    \label{eqn:reward_model}
\end{align}

Note that the reward model in \cref{eqn:reward_model} is non-parametric, i.e. there are 16 unique weights, and $E\left[\reward{t+1}{i}|S_1, S_2, S_3, a\right]$ has 16 dimensions. We follow the methods described in \cite{peng2019} to form our priors using the SARA dataset, which involved fitting linear models like Equation \ref{eqn:reward_model} using GEE. We do a complete-case analysis on the SARA data, and transform it into State-Action-Reward tuples to mimic our RL algorithm setup. However, as noted in the previous section, the SARA dataset does not account for the entire state-space, specifically $S_2$, i.e. time of day, as the participants in the study were requested to self-report on a daily survey just once a day. To that end, we omit all terms from Equation \ref{eqn:reward_model} which contain $S_2$ when forming our analysis to determine priors. Hence, our reward estimation model while forming priors is given as:
\begin{align}
    E\left[\reward{t+1}{i}|S_1,  S_3, a\right] &= \alpha_0 + \alpha_1 S_1 +  \alpha_3 S_3 +  \alpha_5 S_1 S_3 
    + a  (\beta_0 + \beta_1 S_1 + \beta_3 S_3  + \beta_5 S_1 S_3 ).
    \label{eqn:reward_model_noS2}
\end{align}

\subsubsection{State formation and imputation}
We operationalize the states of the RL algorithm from the SARA dataset as follows:
\begin{itemize}
    \item $S_1$: Same as defined in Section \ref{sec:rl_framework}, using  \cref{eqn:S1}
    \item $S_2$: This is not present in the SARA data set.
    \item $S_3$: Same as defined in Section \ref{sec:rl_framework}, using  \cref{eqn:S3}
\end{itemize}
We work with complete-case data. So, whenever there is missing ``cannabis use'' in the past decision point (since $Y=1$ only), we impute and set $S_3$ to 1. This is because participants in the MiWaves study are expected to often use cannabis (at least 3 times a week). Also, since the data is complete-case, we do not use the first day's (i.e. first data point for a participant) reward to fit the model, as we choose to not impute the state of the participant on the first day while forming our priors.

\subsubsection{Feature significance}
\begin{figure}[h]
    \centering
    \includegraphics[scale=0.3]{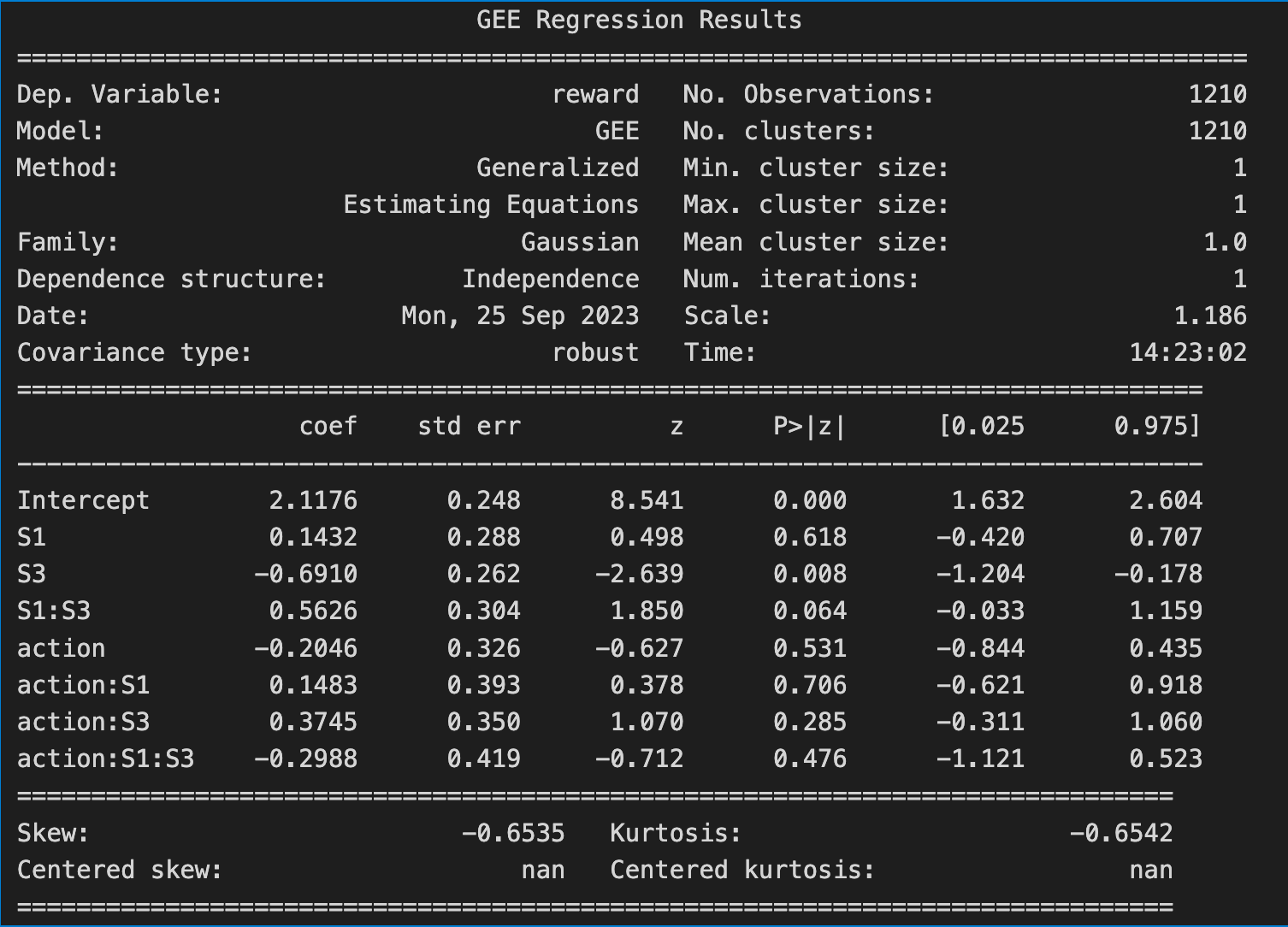}
    \caption{GEE Results}
    \label{fig:gee_analysis}
\end{figure}

We run a GEE linear regression analysis with robust standard errors to determine the feature significance. We categorize a feature to be \emph{significant} when it's corresponding p-value is less than $0.05$. The GEE Regression results are summarized in \Cref{fig:gee_analysis}. Using the criteria mentioned above, we classify the intercept ($\alpha_0$), and the $S_3$ term ($\alpha_3$) to be significant.

\begin{table}[!h]
    \centering
    \begin{tabular}{c|c|c|c}
        Parameter & Significance (S/I/M) & Mean & Variance \\
        \hline
        \hline
        Intercept & S & $2.12$ & $(0.78)^2$\\
        \hline
        $S_1$ & I & $0.00$ & $(0.38)^2$\\
        \hline
        $S_2$ & M & $0.00$ & $(0.62)^2$\\
        \hline
        $S_3$ & S & $-0.69$ & $(0.98)^2$\\
        \hline
        $S_1 S_2$ & M & $0.00$ & $(0.16)^2$\\
        \hline
        $S_1 S_3$ & I & $0.00$ & $(0.1)^2$\\
        \hline
        $S_2 S_3$ & M & $0.00$ & $(0.16)^2$\\
        \hline
        $S_1 S_2 S_3$ & M & $0.00$ & $(0.1)^2$\\
        \hline
        $a$ Intercept & I & $0.00$ & $(0.27)^2$\\
        \hline
        $a S_1$ & I & $0.00$ & ($0.33)^2$\\
        \hline
        $a S_2$ & M & $0.00$ & $(0.3)^2$\\
        \hline
        $a S_3$ & I & $0.00$ & $(0.32)^2$\\
        \hline
        $a S_1 S_2$ & M & $0.00$ & $(0.1)^2$\\
        \hline
        $a S_1 S_3$ & I & $0.00$ & ($0.1)^2$\\
        \hline
        $a S_2 S_3$ & M & $0.00$ & $(0.1)^2$\\
        \hline
        $a S_1 S_2 S_3$ & M & $0.00$ & $(0.1)^2$\\
        \hline
        
    \end{tabular}
    \caption{\textbf{Prior values for the RL algorithm informed using the SARA data set.} The significant column signifies the significant terms found during the analysis using S, insignificant terms using I, and missing terms with M. Values are rounded to the nearest 2 decimal places.}
    \label{tab:informative_prior_vals}
\end{table}

\subsubsection{Initial value of Noise variance (\texorpdfstring{$\sigma^2_{\epsilon}$}{sigma\^2})}
To form an initial of the noise variance, we fit a GEE linear regression model (Equation \ref{eqn:reward_model_noS2}) per participant, and compute the residuals. We set the initial noise variance value as the average of the variance of the residuals across the $N=42$ participant models; that is $\sigma^2_{\epsilon, 0} = 0.85$ in our simulations.

\begin{table}[!h]
    \centering
    \begin{tabular}{c|c}
        Parameter & Value \\
        \hline
        \hline
        $\sigma_{\epsilon, 0}^2$ & $0.85$\\
        \hline
        
    \end{tabular}
    \caption{Initial value of noise variance}
    \label{tab:initial_value_noise_var}
\end{table}

\subsubsection{Prior mean for \texorpdfstring{$\alpha$}{alpha} and \texorpdfstring{$\beta$}{beta}}
To compute the prior means, we first fit a single GEE regression model (Equation \ref{eqn:reward_model_noS2}) across all the participants combined. For the significant features (intercept and $S_3$), we choose the point estimates of $\alpha_0$, and $\alpha_3$ to be their corresponding feature prior means. For the insignificant features, we set the prior mean to 0 ($\alpha_1$, $\alpha_5$, $\beta_0$, $\beta_1$, $\beta_3$, and $\beta_5$). For the prior means of the weights on the $S_2$ terms which are not present in the GEE model, we also set them to 0. 

\subsubsection{Prior standard deviation for \texorpdfstring{$\alpha$}{alpha} and \texorpdfstring{$\beta$}{beta}}
To compute the prior standard deviation, we first fit participant-specific GEE regression models (Equation \ref{eqn:reward_model_noS2}), one per participant. We choose the standard deviation of significant features, corresponding to $\alpha_0$, and $\alpha_3$, across the $N=42$ participant models to be their corresponding prior standard deviations. For the insignificant features, we choose the standard deviation of $\alpha_1$, $\alpha_5$, $\beta_0$, $\beta_1$, $\beta_3$, and $\beta_5$ across the $N=42$ participant models, and set the prior standard deviation to be half of their corresponding values.  The rationale behind setting the mean to 0 and shrinking the prior standard deviation is to ensure stability in the algorithm; we do not expect these terms to play a significant impact on the action selection, unless there is a strong signal in the participant data during the trial. In other words, we are reducing the SD of the non-significant weights because we want to provide more shrinkage to the prior mean of 0 (i.e. more data is needed to overcome the prior). For the prior standard deviation of the $S_2$ terms which are in the baseline (i.e. all the $\alpha$ terms), we set it to the average of the prior standard deviations of the other baseline $\alpha$ terms. Similarly, we set the prior standard deviation of the $S_2$ advantage ($\beta$) terms as the average of the prior standard deviations of the other advantage $\beta$ terms. We further shrink the standard deviation of all the two-way interactions by a factor of 4, and the three-way interactions by a factor of 8 - with a minimum prior standard deviation of $0.1$. Recall that we expect little to no signal from the data wrt. the higher order interactions of the binary variables, thus this decision. Unless there is strong evidence in the participant data during the trial, we do not expect these interaction terms to play a significant role in action selection \cite{gelman2008weakly}.

\subsubsection{Initial values for variance of random effects \texorpdfstring{$\Sigma_u$}{Sigma\_u}}
The mixed effects variant of the algorithm (see \ref{sec:random_effects_alg_variant}) assumes that each weight $\alpha_{i, j}$ is split into a population term $\alpha_{i, \text{pop}}$ and an individual term $u_{i, j}$ (random effects) for a given feature $i$ and participant $j$. These random effects try to capture the participant heterogeneity in the study population, and the variance of the random effects are denoted by $\Sigma_u$. We set the initial values for $\Sigma_u$ to be $\bs{\Sigma}_{u, 0} = (0.1)^2 \times \bs{I}_{K}$, where $K$ is the number of random effects terms. We shrink this variance as we allow the data to provide evidence for participant heterogeneity. We set the off-diagonal entries in the initial covariance matrix to 0 as we do not have any information about the covariance between the random effects.

\subsubsection{Empirical check for prior variance shrinkage}
\label{sec:empirical_check}
While constructing the RL algorithm priors, we shrink the prior variance terms for insignificant, missing and higher order interaction terms (refer \Cref{sec:priors} and \Cref{tab:informative_prior_vals}). This shrinkage allows us to deploy complex models without many drawbacks. However, it relies on the idea that if there is evidence of any such term in the study/data, the algorithm will learn and identify that signal. This empirical check is to establish that our algorithms are able to do so, when such evidence exists. In this empirical check, we are primarily concerned with the variance shrinkage of the action intercept or the advantage intercept coefficient.

\begin{figure}[h]
    \begin{subfigure}[b]{0.45\textwidth}
         \centering
         \includegraphics[width=\textwidth]{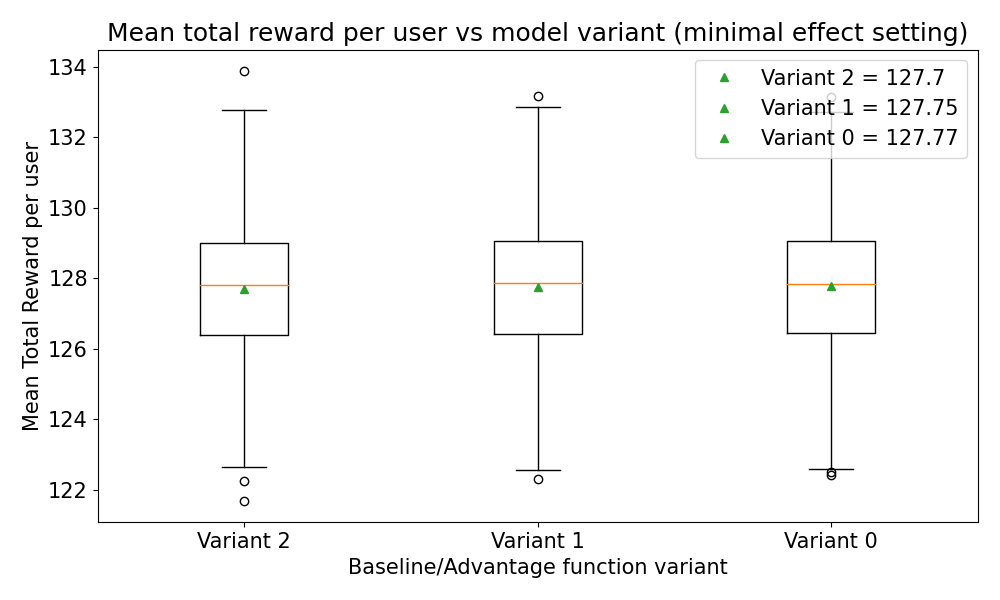}
         \caption{No added signal}
         \label{fig:boxplot}
     \end{subfigure}
     \hfill
     \begin{subfigure}[b]{0.45\textwidth}
         \centering
         \includegraphics[width=\textwidth]{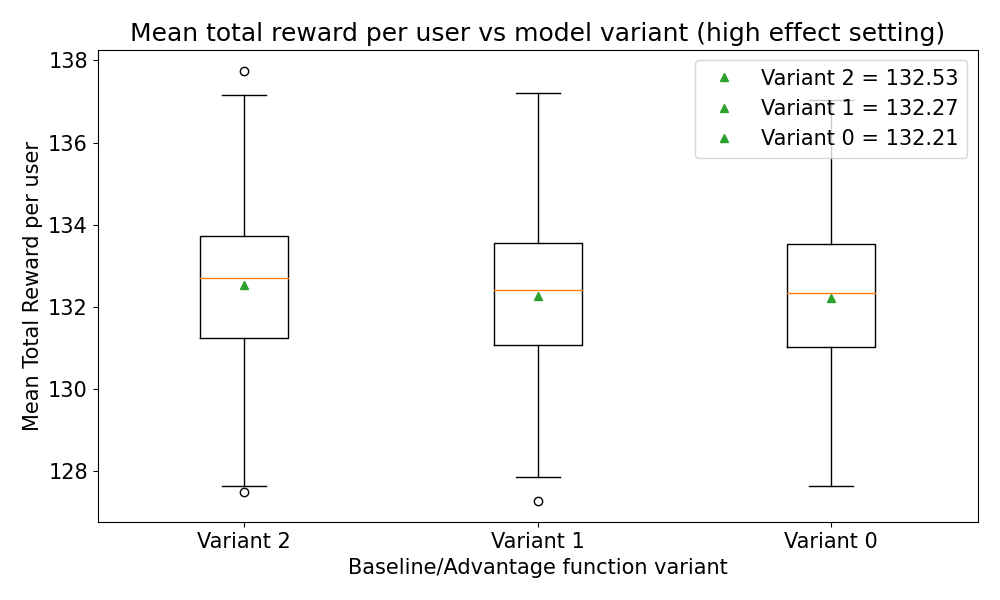}
         \caption{When signal is added to the \emph{action intercept}}
         \label{fig:boxplot_act_int}
     \end{subfigure}
    \caption{Boxplot of Average reward per participant across 500 simulations. $120$ participants per simulation for $T=60$ decision times.}
\end{figure}

\begin{figure}
     \centering
     \begin{subfigure}[b]{0.45\textwidth}
         \centering
         \includegraphics[width=\textwidth]{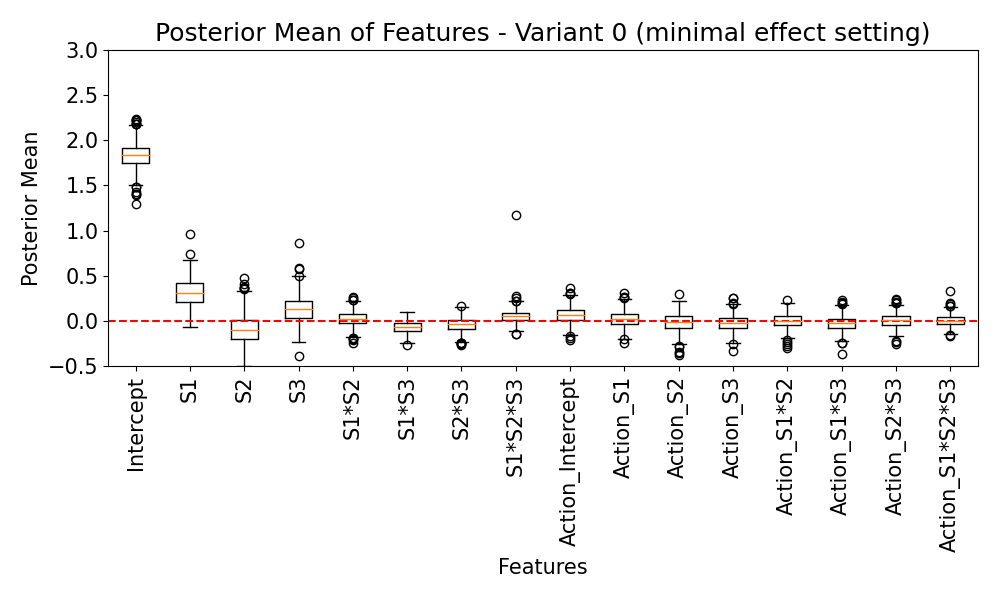}
         \caption{Variant 0: Posterior means}
         \label{fig:post_mean_var0}
     \end{subfigure}
     \hfill
     \begin{subfigure}[b]{0.45\textwidth}
         \centering
         \includegraphics[width=\textwidth]{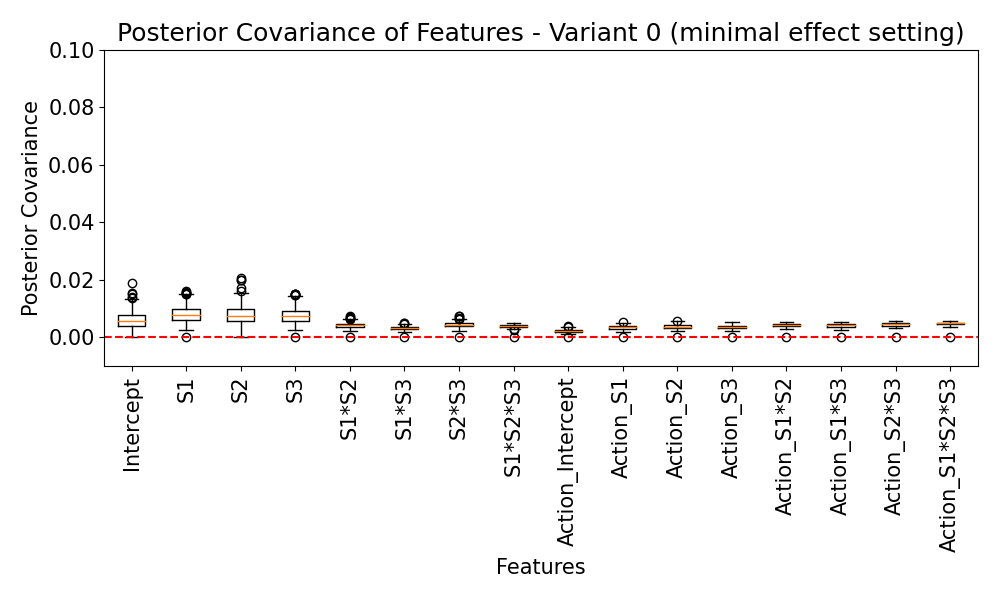}
         \caption{Variant 0: Posterior variance}
         \label{fig:post_cov_var0}
     \end{subfigure}
     \begin{subfigure}[b]{0.45\textwidth}
         \centering
         \includegraphics[width=\textwidth]{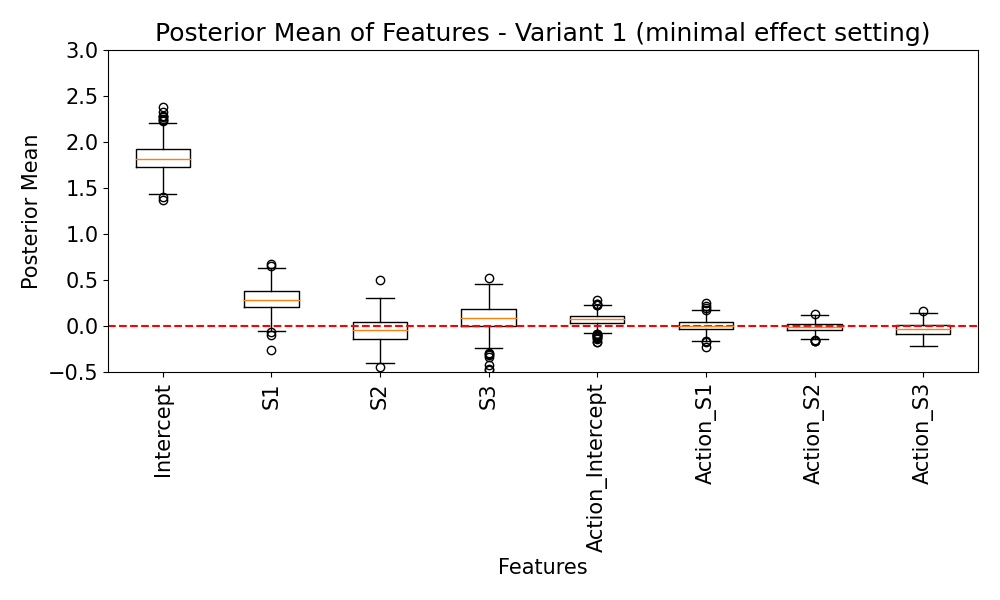}
         \caption{Variant 1: Posterior means}
         \label{fig:post_mean_var1}
     \end{subfigure}
     \hfill
     \begin{subfigure}[b]{0.45\textwidth}
         \centering
         \includegraphics[width=\textwidth]{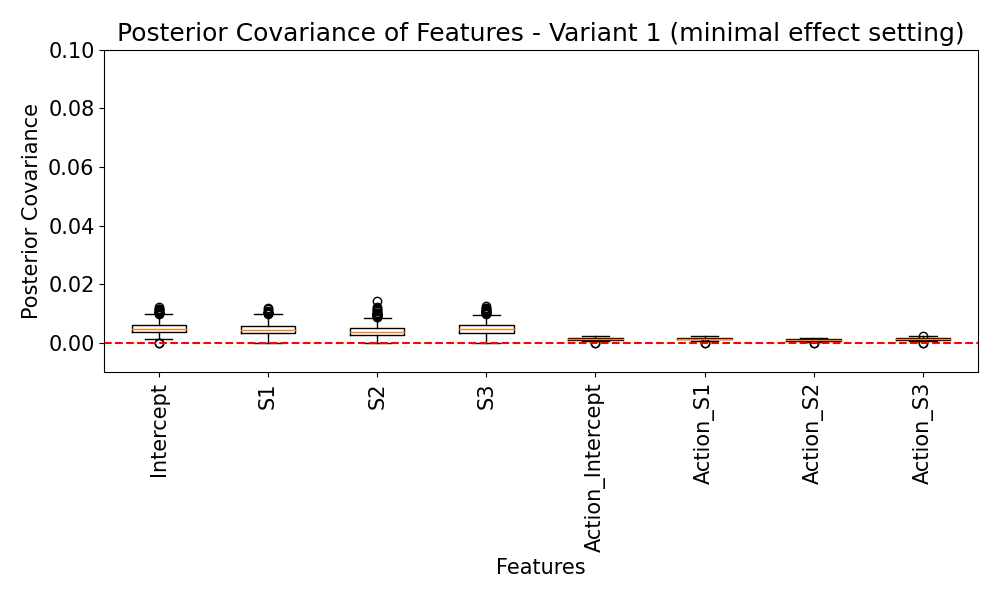}
         \caption{Variant 1: Posterior variance}
         \label{fig:post_cov_var1}
     \end{subfigure}
     \begin{subfigure}[b]{0.45\textwidth}
         \centering
         \includegraphics[width=\textwidth]{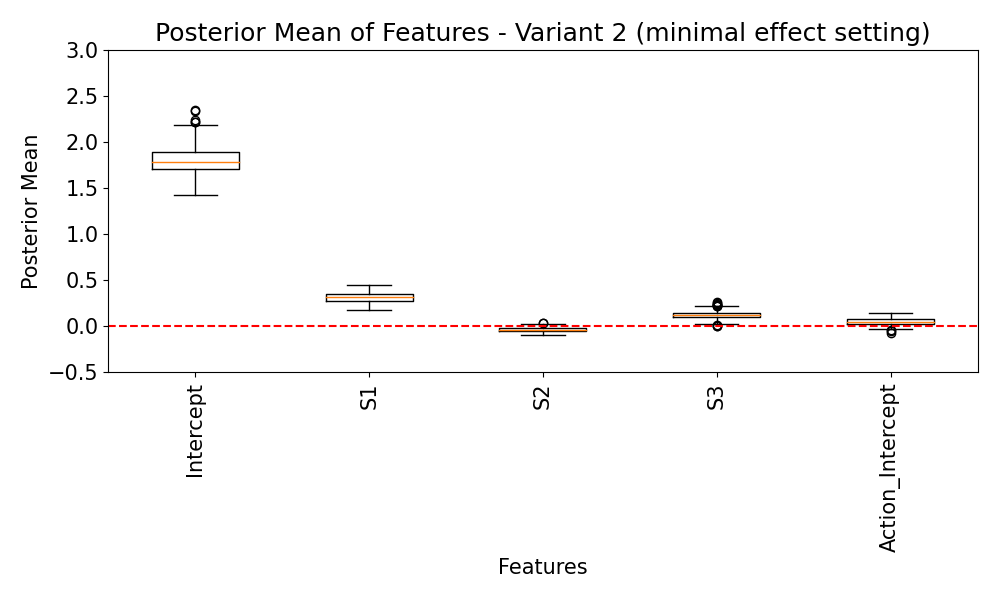}
         \caption{Variant 2: Posterior means}
         \label{fig:post_mean_var2}
     \end{subfigure}
     \hfill
     \begin{subfigure}[b]{0.45\textwidth}
         \centering
         \includegraphics[width=\textwidth]{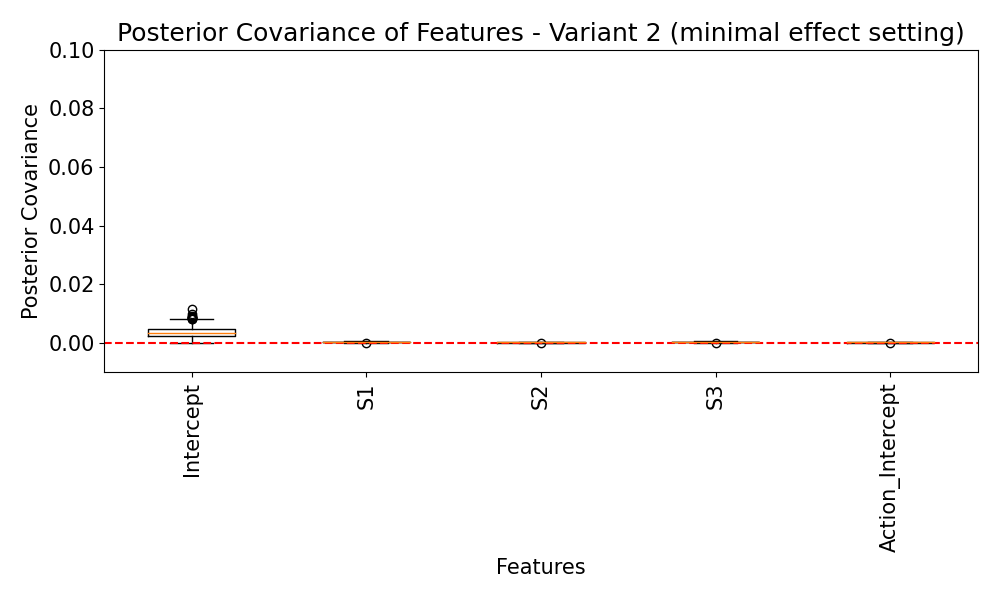}
         \caption{Variant 2: Posterior variance}
         \label{fig:post_cov_var2}
     \end{subfigure}
     \caption{Average posterior means and variances of all the three variants, in the minimal effect setting}
     \label{fig:posteriors_no_signal_pooled_fixed}
\end{figure}

First, we run experiments with no evidence/signal. To that end, we run $500$ simulated MiWaves clinical trials, each trial consisting of $120$ simulated participants and lasting $T=60$ decision times (30 days). We use a smooth clipping function to determine the action selection probability (described in Section \ref{sec:smooth_allocation_function} with $B=20$). We also use the full pooling algorithm (described in Section \ref{sec:random_effects_alg_variant}). The box plot of the average reward per participant across the simulations of all the three baseline and advantage function based variants (described in Section \ref{sec:baseline_adv_func}) can be found in Figure \ref{fig:boxplot}. Their corresponding average posterior means and variances can also be found in Figure \ref{fig:posteriors_no_signal_pooled_fixed}.

Next, we run experiments with evidence/signal in the data, in order to check whether we have shrunk the prior variances too much. To that end, we introduce some signal into the action intercept weight, and check if our pooled algorithm is able to identify and learn that the action weight is nonzero.  To that end, we simulate data with  a \emph{High} treatment effect size, as described in Section \ref{sec:env_variants}, point 4.

\begin{figure}
     \centering
     \begin{subfigure}[b]{0.45\textwidth}
         \centering
         \includegraphics[width=\textwidth]{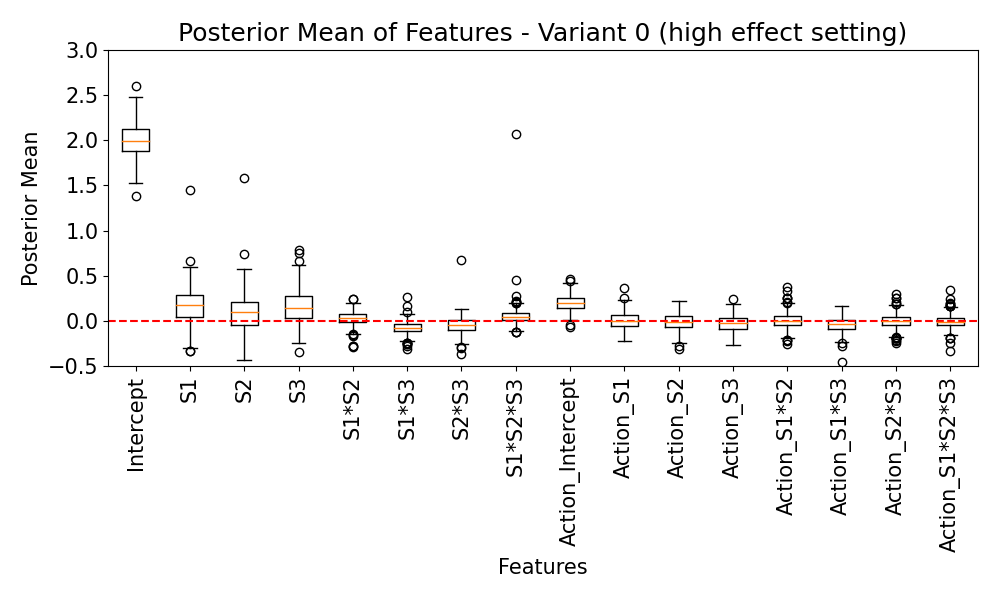}
         \caption{Variant 0: Posterior means}
         \label{fig:post_mean_var0_act_int}
     \end{subfigure}
     \hfill
     \begin{subfigure}[b]{0.45\textwidth}
         \centering
         \includegraphics[width=\textwidth]{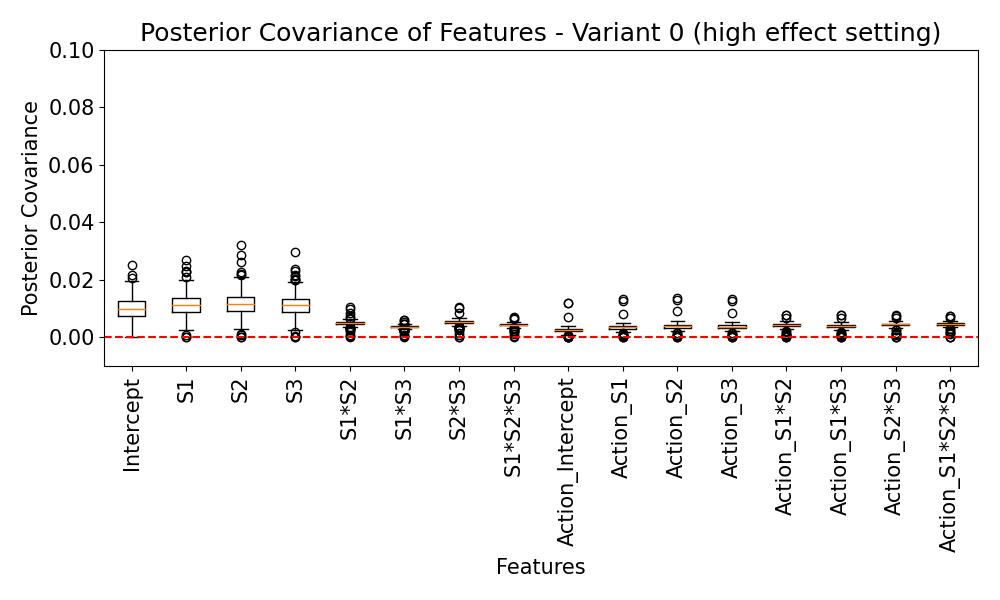}
         \caption{Variant 0: Posterior variance}
         \label{fig:post_cov_var0_act_int}
     \end{subfigure}
     \begin{subfigure}[b]{0.45\textwidth}
         \centering
         \includegraphics[width=\textwidth]{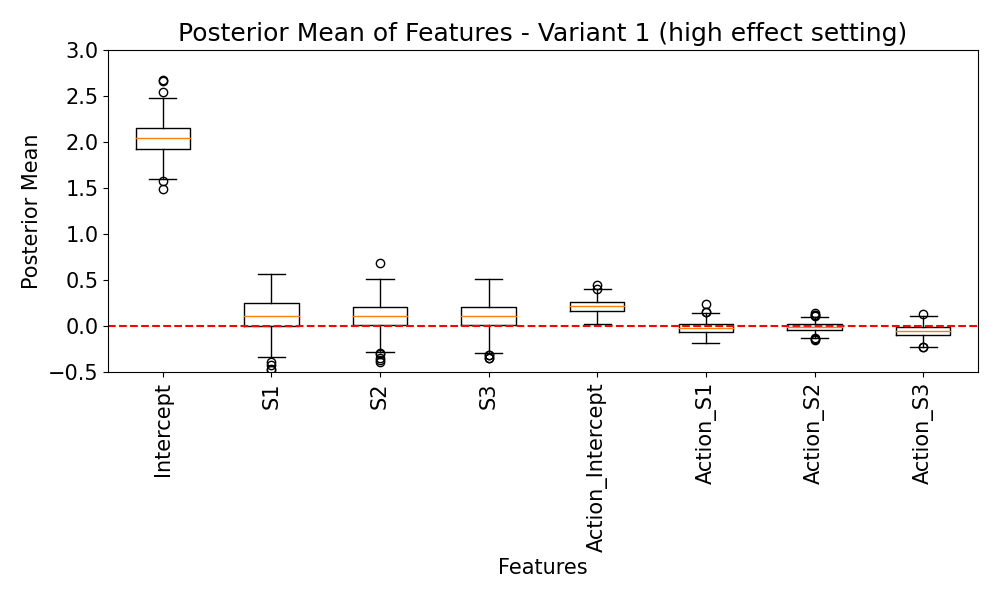}
         \caption{Variant 1: Posterior means}
         \label{fig:post_mean_var1_act_int}
     \end{subfigure}
     \hfill
     \begin{subfigure}[b]{0.45\textwidth}
         \centering
         \includegraphics[width=\textwidth]{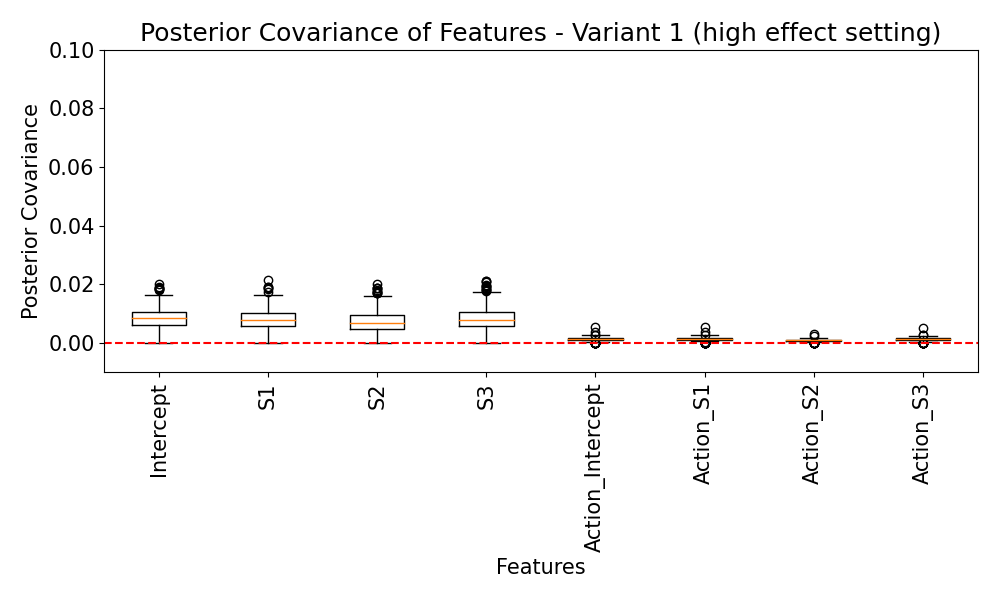}
         \caption{Variant 1: Posterior variance}
         \label{fig:post_cov_var1_act_int}
     \end{subfigure}
     \begin{subfigure}[b]{0.45\textwidth}
         \centering
         \includegraphics[width=\textwidth]{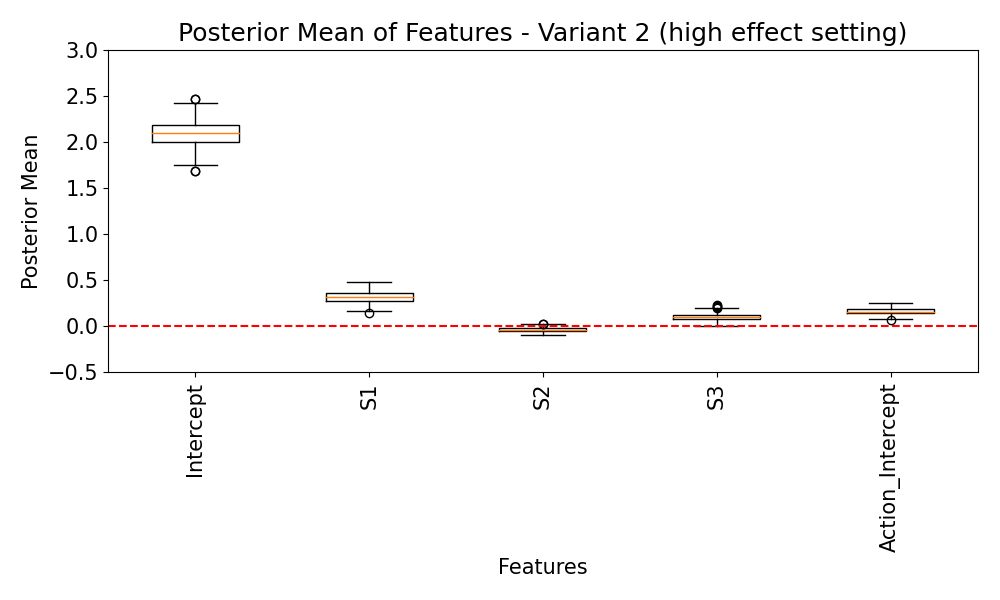}
         \caption{Variant 2: Posterior means}
         \label{fig:post_mean_var2_act_int}
     \end{subfigure}
     \hfill
     \begin{subfigure}[b]{0.45\textwidth}
         \centering
         \includegraphics[width=\textwidth]{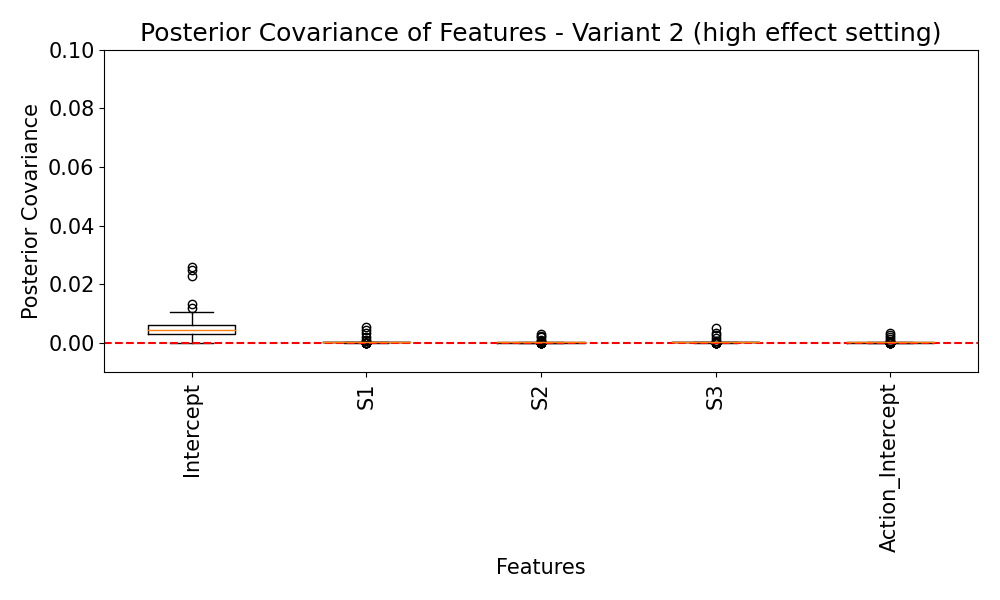}
         \caption{Variant 2: Posterior variance}
         \label{fig:post_cov_var2_act_int}
     \end{subfigure}
     \caption{Average posterior means and variances of all the three variances, when added signal to \emph{action intercept}. It is important to note that the posterior mean of the action intercept is observed to be non-zero. This demonstrates that that the shrinkage of the prior variance of the action intercept term still permits learning when there is strong evidence of a signal.}
     \label{fig:posteriors_act_int_signal_pooled_fixed}
\end{figure}

We again compare the three variants, in terms of the average reward (Figure \ref{fig:boxplot_act_int}), learnt average posterior means and variances (Figure \ref{fig:posteriors_act_int_signal_pooled_fixed}). We see that the algorithm is able to learn the signal added to the action intercept weight. This demonstrates that the shrinkage of the prior variance of the \emph{action intercept} term still permits learning,  when the participant data supplies strong evidence of a signal. \bo{Evidence of the algorithm being able to pick up these signals allow us to run more complex models}. We are able to do so because we shrink the prior variances of insignificant, missing and high order interaction terms (refer \Cref{tab:informative_prior_vals} and \Cref{sec:priors}) by a lot.

\pagebreak

\subsection{Algorithm Variants}
\label{sec:alg_variants}
We will now describe all our algorithm variants in the subsequent sections. We set the lower clipping probability at $L_{\text{min}}=0.2$, while upper clipping probability at $L_{\text{max}} = 0.8$ for all our algorithm variants.

\subsubsection{Fully Pooled vs Mixed effects Model}
\label{sec:random_effects_alg_variant}
We have two variants wrt. the random effects being present in the reward model of our RL algorithm. One is a pooled algorithm, i.e. without \emph{random effects}, while the other is our mixed effects model which utilizes \emph{random effects} (as described in \cref{sec:online_learning_alg}).

\begin{itemize}
    \item \bo{Variant 1: Fully pooled model} \\
    We operationalize the pooled model, across all participants, by fitting the rewards using a single Bayesian Linear Regression model. For a given participant $i$ at decision point $t$, the action-centered training model \cite{greenewald2017action} is defined as:

    \begin{align}
        \reward{t+1}{i} = g(\state{t}{i})^T \bs{\alpha}  + (\action{t}{i} - \pi_i^{(t)}) f(\state{t}{i})^T \bs{\beta} + (\pi_i^{(t)})f(\state{t}{i})^T \bs{\gamma} + \epsilon_i^{(t)}
    \end{align}
    where $\pi_i^{(t)}$ is the probability of taking an active action i.e. sending a MiWaves message ($a_i^{(t)} = 1$). $g(S_i^{(t)})$ and $f(S_i^{(t)})$ are the baseline and advantage vectors respectively. $\epsilon_i^{(t)}$ is the error term, assumed to be independent and identically distributed (i.i.d.) Gaussian noise with mean 0 and variance $\sigma_{\epsilon}^2$.

    The probability $\pi_i^{(t)}$ of sending a MiWaves message ($a_i^{(t)} = 1$) is calculated as:
    \begin{align}
        \pi_i^{(t)} = \mathbb{E}_{{\Tilde{\beta} \sim \mathcal{N}(\mu_{\text{post}}^{(t-1)}, \Sigma_{\text{post}}^{(t-1)})}}[\rho(f(S_i^{(t)})^T \bs{\Tilde{\beta}}) |\mathcal{H}_{1:m_{t-1}}^{(t-1)},  \state{t}{i}]
    \end{align}
    
    where $\mathcal{H}_{i}^{(T)} = \{\state{1}{i}, \action{1}{i}, \reward{2}{i}, \cdots, \state{T}{i}, \action{T}{i}, \reward{T+1}{i}\}$ is the trajectory for a given participant $i$ upto time $t$.
    
    The Bayesian model requires the specification of prior values for $\alpha$, $\beta$ and $\gamma$. While we do not assume the terms $\alpha$, $\beta$ and $\gamma$ to be independent, we have no information about the correlation between the $\alpha$, $\beta$ and $\gamma$ terms, and hence we set the correlation to be 0 in our prior variance $\Sigma_{\text{prior}}$. Hence, ${\bs{\Sigma}}_{\text{prior}} = \text{diag}(\bs{\Sigma_{\alpha}}, \bs{\Sigma_{\beta}}, \bs{\Sigma_{\beta}})$ is the prior variance of all the parameters (note that the posterior var-covariance matrix will have off diagonal elements).  Next ${\bs{\mu}}_{\text{prior}} = (\bs{\mu_{\alpha}}, \bs{\mu_{\beta}}, \bs{\mu_{\beta}})$ is the prior mean of all the parameters.  The prior of each parameter is assumed to be normal and given as:
    \begin{align}
        \bs{\alpha} \sim \mathcal{N}(\bs{\mu_{\alpha}}, \bs{\Sigma_{\alpha}})\\
        \bs{\beta} \sim \mathcal{N}(\bs{\mu_{\beta}}, \bs{\Sigma_{\beta}})\\
        \bs{\gamma} \sim \mathcal{N}(\bs{\mu_{\beta}}, \bs{\Sigma_{\beta}})
    \end{align}

    Since the priors are Gaussian, and the error term is Gaussian, the posterior distribution of all the parameters, given the history of state-action-reward tuples (trajectory) up until time $t$ is also Gaussian. Let us denote all the parameters using $\bs{\theta}^T = (\bs{\alpha}^T, \bs{\beta}^T, \bs{\gamma}^T)$. The posterior distribution of $\bs{\theta}$ given the current history $\mathcal{H}^{(t)}$ = $\{\state{\tau}{}, \action{\tau}{}, \reward{\tau+1}{}\}_{\tau \leq t}$ (data of all participants) is denoted by $\mathcal{N}(\mu_{\text{post}}^{(t)}, \Sigma_{\text{post}}^{(t)})$, where:
    \begin{align}
        \bs{\Sigma}_{\text{post}}^{(t)} &= \bigg( \frac{1}{\sigma_{\epsilon}^2} \bs{\Phi}_{1:t}^T \bs{\Phi}_{1:t} + {\bs{\Sigma}}^{-1}_{\text{prior}} \bigg)^{-1}\\
        \bs{\mu}_{\text{post}}^{(t)} &= \bs{\Sigma}_{\text{post}}^{(t)} \bigg( \frac{1}{\sigma_{\epsilon}^2} \bs{\Phi}_{1:t}^T \mathbf{R}_{2:t+1} + {\bs{\Sigma}}^{-1}_{\text{prior}} {\bs{\mu}}_{\text{prior}} \bigg)
    \end{align}
    
    $\bs{\Phi}_{1:t}$ is a stacked vector of $\bs{\Phi}{}{}(\bs{S}, a)^T = [ g(\state{t}{})^T, (\action{t}{} - \pii{t}{}) f(\state{t}{})^T, \pi^{(t)}f(\state{t}{})^T ]$ for all participants, for all decision points from 1 to $t$. $\mathbf{R}_{1:t}$ is a stacked vector of rewards for all participants, for all decision points from 1 to $t$. $\sigma_{\epsilon}^2$ is the noise variance, ${\bs{\Sigma}}_{\text{prior}} = \text{diag}(\bs{\Sigma_{\alpha}}, \bs{\Sigma_{\beta}}, \bs{\Sigma_{\beta}})$ is the prior variance of all the parameters, and ${\bs{\mu}}_{\text{prior}} = (\bs{\mu_{\alpha}}, \bs{\mu_{\beta}}, \bs{\mu_{\beta}})$ is the prior mean of all the parameters.
    
    We initialize the noise variance by $\sigma_{\epsilon}^2=0.85$; see 
    Equation (\ref{eqn:multistate_optimization_fixed}) for the update equation.  See Section \ref{sec:priors} for the choice of $0.85$. Then we
    update the noise variance by solving the following optimization problem, that is maximizing the marginal (log) likelihood of the observed rewards, marginalized over the parameters:
    \begin{align}
        \sigma_{\epsilon}^2 &= \argmax[- \log(\det(\bs{X} + y \bs{A})) + m_tt \log(y) - y \sum_{\tau\in [t]} \sum_{i\in [m_t]} (R_{i}^{(\tau + 1)})^2 \nonumber \\
        &+ (\bs{X} \bs{\mu_{\text{prior}}}  + y \bs{B})^T (\bs{X} + y \bs{A} )^{-1} (\bs{X} \bs{\mu_{\text{prior}}}  + y \bs{B})]
        \label{eqn:multistate_optimization_fixed}
    \end{align}
    where $\bs{X} = \bs{\Sigma}^{-1}_{\text{prior}}$, $y = \frac{1}{\sigma_{\epsilon}^2}$,  $\bs{A} = \bs{\Phi}_{1:t}^T \bs{\Phi}_{1:t}$, $\bs{B} = \bs{\Phi}_{1:t}^T \mathbf{R}_{2:t+1}$, $t$ is current total number of decision points, and $m_t$ is the number of participants who have been or are currently part of the study.

    \item \bo{Variant 2: Mixed effects model}\\
    This is the same model defined earlier in \cref{rl:reward_model}. Recall that we have the following action-centered \cite{greenewald2017action} training model:
    \begin{align}
        \reward{t+1}{i} = g(\state{t}{i})^T \bs{\alpha_i}  + (\action{t}{i} - \pi_i^{(t)}) f(\state{t}{i})^T \bs{\beta_i} + (\pi_i^{(t)})f(\state{t}{i})^T \bs{\gamma_i} + \epsilon_i^{(t)}
    \end{align}

    Each weight $\bs{\alpha_{i}}$, $\bs{\beta_{i}}$ and $\bs{\gamma_{i}}$ is defined to have a population level term, and an individual level term (random effect). We represent that as follows:
    \begin{align}
        \bs{\alpha_i} = \bs{\alpha_{\text{pop}}} + \bs{u_{\alpha, i}}\\
        \bs{\beta_i} = \bs{\beta_{\text{pop}}} + \bs{u_{\beta, i}}\\
        \bs{\gamma_i} = \bs{\gamma_{\text{pop}}} + \bs{u_{\gamma, i}}
    \end{align}
    where the random effects $\bs{u_i} = [\bs{u_{\alpha, i}}, \bs{u_{\beta, i}}, \bs{u_{\gamma, i}}]^T$ are normal with mean 0 and some variance. That is:   
    \begin{align}
        \bs{u_i} \sim \mathcal{N}(0, \bs{\Sigma_u})
    \end{align}

    We further vectorize the parameters, and represent them as $\bs{\theta}$. We re-write the action-centered reward model as:
    \begin{align}
        \reward{t+1}{i} &= \Phi(\state{t}{i}, \action{t}{i})^T \bs{\theta}_i + \epsilon_{i}^{(t)}\\
        \Phi(\state{t}{i}, \action{t}{i})^T &= [g(\state{t}{i})^T, (\action{t}{i} - \pi_i^{(t)}) f(\state{t}{i})^T, \pi_i^{(t)} f(\state{t}{i})^T]\\
        \bs{\theta}_i &= \begin{bmatrix}
            \bs{\alpha}_i\\
            \bs{\beta}_i\\
            \bs{\gamma}_i
        \end{bmatrix} = \begin{bmatrix}
            \bs{\alpha}_{\text{pop}} + \bs{u}_{\alpha, i}\\
            \bs{\beta}_{\text{pop}} + \bs{u}_{\beta, i}\\
            \bs{\gamma}_{\text{pop}} + \bs{u}_{\gamma, i}
        \end{bmatrix} = \bs{\theta}_{\text{pop}} + \bs{u}_i
    \end{align}

    Overall, we can write, for $m$ participants in the study:
    \begin{align}
        \bs{R} &= \bs{\Phi}(S, a)^T \bs{\theta} +\bs{\epsilon}\\
        \bs{\theta} &= \begin{bmatrix}
            \bs{\theta}_1\\
            \bs{\theta}_2\\
            \vdots\\
            \bs{\theta}_N
        \end{bmatrix} = \begin{bmatrix}
            \bs{\theta}_{\text{pop}} + \bs{u}_1\\
            \bs{\theta}_{\text{pop}} + \bs{u}_2\\
            \vdots\\
            \bs{\theta}_{\text{pop}} + \bs{u}_{m_t}
        \end{bmatrix}\\
        \bs{\epsilon}_i &=
        \begin{bmatrix}
            \epsilon_{i1} \\
            \epsilon_{i2} \\
            \vdots\\
            \epsilon_{it}
        \end{bmatrix}\;\;\;
        \bs{\epsilon}=
        \begin{bmatrix}
            \bs{\epsilon_{1}} \\
            \bs{\epsilon_{2}} \\
            \vdots\\
            \bs{\epsilon_{m_t}}
        \end{bmatrix}\\
        \bs{u}_i &\sim \mathcal{N}(\bs{0}, \bs{\Sigma}_{u})\\
        \bs{\epsilon} &\sim \mathcal{N}(\bs{0}, \sigma_{\epsilon}^2\bs{I}_{tm_t})
    \end{align}

    We assume Gaussian priors on the population level terms $\bs{\theta}_{\text{pop}} \sim \mathcal{N}(\bs{\mu}_{\text{prior}}, \bs{\Sigma}_{\text{prior}})$. While we do not assume the terms $\bs{\alpha}_{\text{pop}}$, $\bs{\beta}_{\text{pop}}$ and $\bs{\gamma}_{\text{pop}}$ to be independent, we have no information about the covariance between the $\bs{\alpha}_{\text{pop}}$, $\bs{\beta}_{\text{pop}}$ and $\bs{\gamma}_{\text{pop}}$ terms, and hence we set the covariance to be 0 in our prior variance $\bs{\Sigma_{\text{prior}}}$. Hence, ${\bs{\Sigma}}_{\text{prior}} = \text{diag}(\bs{\Sigma_{\alpha}}, \bs{\Sigma_{\beta}}, \bs{\Sigma_{\beta}})$ is the prior variance of all the parameters (note that the posterior variance-covariance matrix will have off diagonal elements).  Next ${\bs{\mu}}_{\text{prior}} = (\bs{\mu_{\alpha}}, \bs{\mu_{\beta}}, \bs{\mu_{\beta}})$ is the prior mean of all the parameters. The prior of each parameter is assumed to be normal and given as:
    \begin{align}
        \bs{\alpha}_{\text{pop}} \sim \mathcal{N}(\bs{\mu_{\alpha}}, \bs{\Sigma_{\alpha}})\\
        \bs{\beta}_{\text{pop}} \sim \mathcal{N}(\bs{\mu_{\beta}}, \bs{\Sigma_{\beta}})\\
        \bs{\gamma}_{\text{pop}} \sim \mathcal{N}(\bs{\mu_{\beta}}, \bs{\Sigma_{\beta}})
    \end{align}
    We use the values from Table \ref{tab:informative_prior_vals} to inform both ${\bs{\mu}}_{\text{prior}}$ and ${\bs{\Sigma}}_{\text{prior}}$. Also, since we don't know $\bs{\Sigma_{u}}$, we initialize this matrix at time $t=0$ as $\bs{\Sigma_{u, 0}}$. We do the same for $\sigma^2_{\epsilon}$, and initialize it as $\sigma_{\epsilon, 0}^2=0.85$; see Equation (\ref{eqn:argmaxLL}) for the update equation.  See Section \ref{sec:priors} for the choice of $0.85$. We set the initial value of $\bs{\Sigma}_{u, 0} = (0.1)^2 \times \bs{I}_{K}$ ($K = \text{dim}(\Phi(S_{i}^{(t)}, \action{t}{i}))$). We note that this is highly unusual: the prior variance values in Table \ref{tab:informative_prior_vals} were obtained after fitting a GEE model. These variance values not only exhibit the uncertainty in estimating the weights from the SARA data, but they also capture the heterogeneity amongst the population present in the SARA data. The random effects variance also aims to capture the heterogeneity amongst the study population. However, we allow the prior variances and initial values of random effects variances to be set as mentioned above (we have shrunk them), as want the data to provide evidence of these interaction terms. The rest of the posterior and hyper-parameter update follows from \Cref{rl:update}.

\end{itemize}


\subsubsection{Models for Baseline and Advantage Functions}
\label{sec:baseline_adv_func}
We construct three variants of our reward model, by varying the baseline and advantage features:
\begin{itemize}
    \item \bo{Variant 0:} All interactions in the baseline and advantage:
    \begin{align}
        g(S) &= \{1, S_1, S_2, S_3, S_1 S_2, S_2 S_3, S_1 S_3, S_1 S_2 S_3\}\\
        f(S) &= \{1, S_1, S_2, S_3, S_1 S_2, S_2 S_3, S_1 S_3, S_1 S_2 S_3\}
    \end{align}
    \item \bo{Variant 1:} Only one-way interactions in the baseline and advantage:
    \begin{align}
        g(S) &= \{1, S_1, S_2, S_3\}\\
        f(S) &= \{1, S_1, S_2, S_3\}
    \end{align}
    \item \bo{Variant 2:} Only one-way interactions in the baseline, and only the intercept term in the advantage:
    \begin{align}
        g(S) &= \{1, S_1, S_2, S_3\}\\
        f(S) &= \{1\}
    \end{align}
\end{itemize}

\subsubsection{Smooth Allocation Function}
\label{sec:smooth_allocation_function}
Recall from \cref{sec:act_select} that the probability of sending a MiWaves message is computed as:
\begin{align}
    \pi_i^{(t)} = \mathbb{E}_{{\Tilde{\beta} \sim \mathcal{N}(\mu_{\text{post}}^{(t-1)}, \Sigma_{\text{post}}^{(t-1)})}}[\rho(f(\state{t}{i})^T \bs{\Tilde{\beta}}) |\mathcal{H}_{1:m_{t-1}}^{(t-1)},  \state{t}{i}]
\end{align}
and $\rho$ is generalized logistic function, defined as follows:
\begin{equation}
    \rho(x) = L_{\min} + \frac{ L_{\max} - L_{\min} }{  1 + c \exp(-b x) }
    \label{eqn:smooth_post_sampling}
\end{equation}

where $L_{\min} = 0.2$ and $L_{\max} = 0.8$ are the lower and upper clipping probabilities. We set $c=5$. Larger values of $c > 0$ shifts the value of $\rho(0)$ to the right.  This choice implies that $\rho(0)=0.3$ (intuitively, the probability of taking an action when treatment effect is $0$, is $0.3$).

We construct two variants, based on the values of $b = \frac{B}{\sigma_{\text{res}}}$, described below - 
\begin{itemize}
    \item \bo{Variant 1: $B = 10$}
    \item \bo{Variant 2: $B = 20$}
\end{itemize}

Larger values of $b > 0$ makes the slope of the curve more ``steep''. $\sigma_{\text{res}}$ is the reward residual standard deviation obtained from fitting our reward model on data from SARA. We have $\sigma_{\text{res}}$ in the denominator, as it helps us standardize the treatment effect (intuition - the denominator in $\rho(f(s) \beta)$ has the term $\beta / \sigma_{\text{res}}$ in the exponential, which becomes the standardized treatment effect).

We currently set $\sigma_{\text{res}} = 0.95$. In order to arrive at this value, we first simulated each of $N=42$ unique participants from SARA in a MiWaves simulation with low treatment effect. The reason we introduce low treatment effect, is because we see that SARA itself had minimal to no treatment effect. Actions are selected with $0.5$ probability. We ran 500 such simulations, and fit each simulation's data into a GEE model, and computed the standard deviation of the residuals. $0.95$ was the mean of the residual standard deviation across these 500 simulations.

\subsubsection{Posterior Update Cadence}
We construct two variants based on the cadence of the posterior update:
\begin{itemize}
    \item \bo{Daily update}: The posterior mean and variance are updated once daily at 4 AM.
    \item \bo{Weekly update}: The posterior mean and variance are updated once weekly on Sundays at 4 AM.
\end{itemize}

\subsubsection{Hyper-parameter Update Cadence}
We construct two variants based on the cadence of the hyper-parameter update:
\begin{itemize}
    \item \bo{Daily update}: The random effects variance and the noise variance are updated once daily at 4 AM.
    \item \bo{Weekly update}: The random effects variance and the noise variance are updated weekly on Sundays at 4 AM.
\end{itemize}

\noindent We perform the following checks during the hyper-parameter optimization step:
\begin{itemize}
    \item $\Sig{}{u}$ is positive-definite (PD): We parameterize the PD covariance matrix in the form of $LL^T$, where $L$ is a lower triangular matrix (cholesky decomposition).
    \item $\bs{\Tilde{\Sigma}_{\theta, t}}$ is PD: We check the eigenvalues, and make sure that they are positive.
    \item $\bs{X} + y\bs{A}$ is PD: We check the eigenvalues, and make sure that they are positive.
    \item Check and make sure $\sigma_e^2 > 0$.
\end{itemize}


\subsection{Results \& Decisions}
\label{sec:results}
We run a set of simulations, consisting of $9$ environment variants, and $24$ algorithm variants. Each algorithm variant is run for $K=500$ simulated trials (also referred to as simulations) in a given environment. Each simulated trial has $m=120$ participants, over a period of 30 days ($T=60$ decision times). We then compute the following metrics for each algorithm variant in each simulated trial:

\begin{itemize}
    \item Average total reward per participant, averaged across participants in a simulated trial: \\$\frac{1}{m} \sum_{i=1}^{m}\sum_{t=1}^{T} \reward{t+1}{i}$, where $m=120$ participants and $T=60$ decision times in a simulated trial.
    \item The median total reward amongst all participant's total reward in each simulated trial.
    \item The average total reward per participant for the lower 25 percentile of participants (in terms of their total reward in a given simulation) in a simulated trial.
    \item The median total reward amongst the lower 25 percentile of participants (in terms of their total reward in a given simulation) in a simulated trial.
\end{itemize}

We use these metrics to evaluate algorithm performance not just in the average case, but also for worse-off users. We then compute the mean and standard deviation of the metrics above, across the $K=500$ simulations in each environment. The resulting metrics can be found in \href{https://docs.google.com/spreadsheets/d/1SnxTgslzx_tS0gBOy_EzrbkYf3bz9DRiKCYYM7M9ThM/edit?usp=sharing}{this spreadsheet}. Based on these set of simulations, we make the following observations and algorithm design decisions:
\begin{itemize}
    \item \bo{$B$ in smooth allocation function}: Keeping all other algorithm variant choices to be the same, $B=20$ based variants consistently outperform $B=10$ based variants during our simulated trials. This trend in their relative performance is seen in both the total population vs the lower 25 percentile of the population (in terms of total reward). 


    \bo{Decision}: We have decided to use $B=20$ for our smooth posterior sampling function. In future, we might also consider power for the primary analysis in selecting $B$.

    \item \bo{Baseline and Advantage Functions}: Recall that we have three variants with respect to the choice of the baseline and advantage functions, as specified in Section \ref{sec:baseline_adv_func}. Reiterating, we have:
    \begin{itemize}
        \item \bo{Variant 0:} All interactions in the baseline and advantage:
        \begin{align}
            g(S) &= \{1, S_1, S_2, S_3, S_1 S_2, S_2 S_3, S_1 S_3, S_1 S_2 S_3\}\\
            f(S) &= \{1, S_1, S_2, S_3, S_1 S_2, S_2 S_3, S_1 S_3, S_1 S_2 S_3\}
        \end{align}
        \item \bo{Variant 1:} Only one-way interactions in the baseline and advantage:
        \begin{align}
            g(S) &= \{1, S_1, S_2, S_3\}\\
            f(S) &= \{1, S_1, S_2, S_3\}
        \end{align}
        \item \bo{Variant 2:} Only one-way interactions in the baseline, and only the intercept term in the advantage:
        \begin{align}
            g(S) &= \{1, S_1, S_2, S_3\}\\
            f(S) &= \{1\}
        \end{align}
    \end{itemize}
    During our simulations, we find that Variant $2$ consistently performs better than Variant $0$ and Variant $1$. However, this is expected; Variant $2$ has an advantage due to the way our simulation environments have been constructed. To elaborate further,  only  the action intercept weight is nonzero in our generative participant models used in the simulation environments. Variant $2$ best explains the observed rewards in such environments since it has only the action intercept in the advantage function. Hence, we observe that Variant $2$ to be performing consistently better in our simulations.  

    However, the entire purpose of having Variants $0$ and $1$ is to observe the cost of incorporating a more complex model. And to that end, we observe that there is very little price to be paid in terms of performance when using such variants.
    

    \bo{Decision}: Use the baseline and advantage functions from Variant $0$ for the RL algorithm.
    
    \item \bo{Fully pooled vs Mixed Effects}: Keeping all other algorithm variant choices to be the same, we find that mixed effects based reward model variants perform better than fully pooled reward model variants. This trend in their relative performance is seen in both the total population vs the lower 25 percentile of the population (in terms of total reward). When using smooth posterior sampling, i.e. when $B < \infty$, algorithms are forced to explore. The fully pooled variants are not able to pick up the heterogeneity amongst participants. On the other hand, the mixed effects models explore and are able to explain the heterogeneity amongst participants, while also learning the reward model of the participants (and any corresponding treatment effect).

    \bo{Decision}: We decided that we will use mixed effects reward model for the RL algorithm - and empirically showed that it performs better than the fixed effects reward model.

    \item \bo{Update cadence}: Keeping all other algorithm variant choices to be the same, we see that when both the posterior update and the hyper-parameter update are done at a weekly basis, the RL algorithm performs relatively worse in comparison to when they are either both updated daily, or the former is updated daily and the latter updated weekly. This trend is observed across all environments, and across all metrics mentioned above.

    As for the performance between the cases where both are updating daily vs the case where posteriors are updated daily and hyper-parameters are updated weekly, there is no clear best performer.

    \bo{Decision}: We decided that we would be updating the posteriors daily and the hyper-parameters weekly going forward.

\end{itemize}

Incorporating all the decisions above, the RL algorithm for the MiWaves study consisting $m$ participants is described in \Cref{algo:multistate_mixed_algo}.

\begin{algorithm}[H]
    \caption{MiWaves RL Algorithm}
    \label{algo:multistate_mixed_algo}
    \SetKwInOut{Input}{Input}
    \SetKwInOut{Output}{Output}
    \SetKwData{n}{$D$}
    \SetKwData{d}{$d$}
    \SetKwData{t}{$\tau$}
    \SetKwData{i}{$i$}
    \SetKwData{j}{$j$}
    \SetKwData{priormeanMu}{$\bs{\mu_0}$}
    \SetKwData{priorcovMu}{$\bs{\Sigma_0}$}
    \SetKwData{priorcovB}{$\bs{\Sigma_{u, 0}}$}
    \SetKwData{priorcovE}{$\sigma^2_{\epsilon, 0}$}
    \SetKwData{priormeanTh}{$\bs{\mu_{\theta}}$}
    \SetKwData{priorcovTh}{$\bs{\Sigma_{\theta}}$}
    \SetKwData{OldmeanThti}{$\bs{\mu^{(\d-1)}_{\text{post}, i}}$}
    \SetKwData{OldcovThti}{$\bs{\Sigma^{(\d-1)}_{\text{post}, i}}$}
    \SetKwData{OldmeanTht}{$\bs{\mu^{(\d-1)}_{\text{post}}}$}
    \SetKwData{OldcovTht}{$\bs{\Sigma^{(\d-1)}_{\text{post}}}$}
    \SetKwData{OldcovBt}{$\bs{\Sigma_{u, \d-1}}$}
    \SetKwData{OldcovEt}{$\sigma^2_{\epsilon, \d-1}$}
    \SetKwData{meanTht}{$\bs{\mu^{(\d)}_{\text{post}}}$}
    \SetKwData{covTht}{$\bs{\Sigma^{(\d)}_{\text{post}}}$}
    \SetKwData{covBt}{$\bs{\Sigma_{u, \t}}$}
    \SetKwData{covEt}{$\sigma^2_{\epsilon, \t}$}
    \SetKwData{Actit}{$\pi^{(\t)}_{i}$}
    \SetKwData{Rho}{$\rho$}
    \Input{\n \ -- Number of days. \\
            $m_t$ -- Number of participants who have been or are part of the study at time $t$. \\
            $\bs{\mu^{(0)}_{\text{post}}} = \muprior$, $\bs{\Sigma^{(0)}_{\text{post}}} = \Sigprior$ \ -- Prior mean and variance of population term, as described in \cref{sec:priors}. \\
            \priorcovB, \priorcovE \ -- Initial values of random effects variance and noise variance respectively, as described in \cref{sec:priors} and \cref{sec:random_effects_alg_variant}.\\
            $\Rho(x)$ -- Smoothing function for posterior sampling from \cref{eqn:smooth_post_sampling}
            }
    \For{\texttt{day }\d $=1$ \KwTo \n}
    {
        \For{\texttt{time of day }\j $=0$ \KwTo 1}
        {
            Compute timestep $\t = (d \times 2) + j$ \\
            \For{\i $=1$ \KwTo $m_t$}
            {
                Observe state $\state{\t}{i}$\;
                Get posteriors $\OldmeanThti$ and $\OldcovThti$ for participant \i from \OldmeanTht and \OldcovTht \;
                Compute action selection probability \Actit:
                \begin{align}
                    \Actit = \E_{\Tilde{w} \sim \mathcal{N}(\OldmeanThti, \OldcovThti)} \bigg[ \rho((\Phii{}{}(s, 1) - \Phii{}{}(s, 0))^T \Tilde{w}) \big| \history{1:m_{\t-1}}{(\t-1)}, \state{\t}{i} \bigg] \nonumber
                \end{align}\\
                Sample action $\action{t}{i} = {\rm Bern} (\Actit)$\\
                Collect reward $R^{(\t + 1)}_{i}$ 
            }
        }
        When $\d \mod 7 = 0$, update $\Sig{}{u, \d}$ and $\sigma^2_{\epsilon, \d}$ using \cref{eqn:argmaxLL} \;
        Update posteriors \meanTht and \covTht using \cref{eqn:postMeanTheta} and \cref{eqn:postCovTheta}
    }
\end{algorithm}